\documentclass[journal]{IEEEtran}
\ifCLASSINFOpdf
\else
\fi
\usepackage{amsmath}
\usepackage[pdftex]{graphicx}
\usepackage{times}
\usepackage{caption2}

\usepackage{amsmath}
\usepackage{amsfonts}
\usepackage{amssymb}
\usepackage[mathscr]{euscript}

\usepackage{bm}
\usepackage{subfigure}
\usepackage{algorithm}
\usepackage{algorithmic}
\usepackage{multirow}
\usepackage{wrapfig}
\usepackage{booktabs}
\usepackage{threeparttable}
\usepackage[english]{babel}
\usepackage{microtype}

\newcommand{\bfC}{\mathbf{C}}
\newcommand{\bfE}{\mathbf{E}}
\newcommand{\bfe}{\mathbf{e}}
\newcommand{\bfF}{\mathbf{F}}

\newcommand{\bft}{\mathbf{t}}
\newcommand{\bfR}{\mathbf{R}}
\newcommand{\bfw}{\mathbf{w}}
\newcommand{\bfU}{\mathbf{U}}

\newcommand{\bfn}{\mathbf{n}}

\newcommand{\bfd}{\mathbf{d}}

\newcommand{\bfr}{\mathbf{r}}

\newcommand{\bfX}{\mathbf{X}}

\newcommand{\projdirection}{\mathbf{Z}}

\newcommand{\bfD}{\mathbf{D}}

\newcommand{\bfv}{\mathbf{v}}
\newcommand{\bfH}{\mathbf{H}}

\newcommand{\wexp}{\mathbf{w}_{\textrm{exp}}}
\newcommand{\wid}{\mathbf{w}_{\textrm{id}}}
\newcommand{\widc}[1]{w_{\textrm{id}}^{(#1)}}
\newcommand{\didc}[1]{\delta_{\textrm{id}}^{(#1)}}
\newcommand{\wexpc}[1]{w_{\textrm{exp}}^{(#1)}}
\newcommand{\dexpc}[1]{\delta_{\textrm{exp}}^{(#1)}}
\newcommand{\efit}{E_{\textrm{fit}}}
\newcommand{\eshade}{E_{\textrm{shading}}}
\newcommand{\proj}{\bm{\Pi}}
\newcommand{\weightfit}[1]{\gamma_{#1}}
\newcommand{\weightalbedo}[1]{\mu_{#1}}
\newcommand{\drefc}[1]{\delta_{\textrm{r}}^{(#1)}}
\newcommand{\weightsubspace}{\mu_2}

\newcommand{\sphcoef}{\bm{\xi}}
\newcommand{\graphlap}{\mathbf{K}^{i}}
\newcommand{\faceregion}{\Omega}
\newcommand{\normalmap}[1]{\mathbf{n}'_{#1}}
\newcommand{\heightfieldnormal}[1]{\widehat{\mathbf{n}}_{#1}}
\newcommand{\egrad}{E_{\textrm{grad}}}
\newcommand{\eclose}{E_{\textrm{close}}}
\newcommand{\esmooth}{E_{\textrm{smooth}}}
\newcommand{\eint}{E_{\textrm{int}}}
\newcommand{\weightnormalmap}[1]{\omega_{#1}}

\newcommand{\ceres}{\textsc{Ceres}}

\newcommand{\facewarehouse}{\mbox{\textsc{FaceWarehouse}}}
\newcommand{\bfm}{\mbox{\textsc{BFM2009}}}

\newcommand{\paraheading}[1]{\textbf{{#1}.~}}

\usepackage[bookmarks,backref=true,linkcolor=black]{hyperref}
\hypersetup{
  pdfauthor = {},
  pdftitle = {},
  pdfsubject = {},
  pdfkeywords = {},
  colorlinks=true,
  linkcolor= black,
  citecolor= black,
  pageanchor=true,
  urlcolor = black,
  plainpages = false,
  linktocpage
}
\usepackage{color}

\definecolor{turquoise}{cmyk}{0.65,0,0.1,0.1}
\definecolor{purple}{rgb}{0.65,0,0.65}
\definecolor{dark_green}{rgb}{0, 0.5, 0}
\definecolor{orange}{rgb}{0.8, 0.2, 0.2}

\newcommand{\revise}[1]{{\color{black}#1}}

\begin{document}
\title{3D Face Reconstruction with Geometry Details from a Single Image}


\author{Luo~Jiang,   Juyong~Zhang$^\dagger$, Bailin~Deng,~\IEEEmembership{Member,~IEEE,}  Hao Li, and~Ligang~Liu,~\IEEEmembership{Member,~IEEE,}% <-this % stops a space
\IEEEcompsocitemizethanks{\IEEEcompsocthanksitem L. Jiang, J. Zhang, H. Li, and L. Liu are with School of Mathematical Sciences,
	University of Science and Technology of China.\protect\\
\IEEEcompsocthanksitem B. Deng is with  School of Computer Science and Informatics, Cardiff University.}% <-this % stops an unwanted space
\thanks{$^\dagger$Corresponding author. Email: \href{mailto:juyong@ustc.edu.cn}{\texttt{juyong@ustc.edu.cn}}.}
}

%\author{\IEEEauthorblockN{Hao Li\IEEEauthorrefmark{1},
%Juyong Zhang\IEEEauthorrefmark{1}\thanks{Corresponding author: %Juyong Zhang (email: juyong@ustc.edu.cn).},
%Jianmin Zheng\IEEEauthorrefmark{2},
%Jianfei Cai\IEEEauthorrefmark{2}}
%\IEEEauthorblockA{\IEEEauthorrefmark{1}School of Mathematical %Sciences,
%University of Science and Technology of China}
%\IEEEauthorblockA{\IEEEauthorrefmark{2}School of Computer %Engineering,
%Nanyang Technological University}}

%\markboth{~}{~}
\markboth{Submitted to IEEE Transactions on Image Processing}
{Jiang \MakeLowercase{\textit{et al.}}: 3D Face Reconstruction with Geometry Details from a Single Image}



% for Transactions on Magnetics papers, we must declare the abstract and
% index terms PRIOR to the title within the \IEEEtitleabstractindextext
% IEEEtran command as these need to go into the title area created by
\IEEEtitleabstractindextext{%
\begin{abstract}
3D face reconstruction from a single image is a classical and challenging problem, with wide applications in many areas. Inspired by recent works in face animation from RGB-D or monocular video inputs, we develop a novel method for reconstructing 3D faces from unconstrained 2D images, using a coarse-to-fine optimization strategy. First, a smooth coarse 3D face is generated from an example-based bilinear face model,  by aligning the projection of 3D face landmarks with 2D landmarks detected from the input image. Afterwards, using local corrective deformation fields, the coarse 3D face is refined using photometric consistency constraints, resulting in a medium face shape. Finally, a shape-from-shading method is applied on the medium face to recover fine geometric details. Our method outperforms state-of-the-art approaches in terms of accuracy and detail recovery, which is demonstrated in extensive experiments using real world models and publicly available datasets.
\end{abstract}

}

\maketitle

\IEEEdisplaynontitleabstractindextext

% \fi
\IEEEpeerreviewmaketitle

\section{Introduction}
Reconstruction of 3D face models using 2D images is a fundamental problem in computer vision and graphics~\cite{Stylianou2009}, with various applications such as face recognition~\cite{BlanzV03,ZhuLYYL15} and animation~\cite{CaoHZ14,Thies_2016_CVPR}. However, this problem is particularly challenging, due to the loss of information during camera projection.
 
In the past, a number of methods have been proposed for face construction using a single image. Among them, example-based methods first build a low-dimensional parametric representation of 3D face models from an example set, and then fit the parametric model to the input 2D image. One of the most well-known examples is the 3D Morphable Model (3DMM) proposed by Blanz and Vetter~\cite{BlanzV99}, represented as linear combination of the example faces. 3DMM is a popular parametric face model due to its simplicity, and has been the foundation of other more sophisticated face reconstruction methods~\cite{ZhuLYYL15}. Another approach to single image reconstruction is to solve it as \emph{Shape-from-shading} (SFS)~\cite{Zhang99shapefrom}, a classical computer vision problem of 3D shape recovery from shading variation. For example, Kemelmacher-Shlizerman and Basri~\cite{Kemelmacher-ShlizermanB11} reconstruct the depth information from an input face image, by estimating its lighting and reflectance parameters using a reference face shape.

While these existing approaches are able to produce high-quality reconstruction from a single image, they also come with limitations. Although example-based methods are simple and efficient, they rely heavily on the dataset, and may produce unsatisfactory results when the target face is largely different from those in the example set; moreover, due to the limited degrees of freedom of the low-dimensional model, these methods often fail to reproduce fine geometric details (such as wrinkles) that are specific to the target face. SFS-based methods are able to capture the fine-scale facial details from the appearance of the input image; however, they require prior knowledge about the geometry or illumination to resolve the ambiguity of the reconstruction problem, and may become inaccurate when the input image does not satisfy the assumptions. 

\begin{figure}[t!] 
	\centering
	\includegraphics[width=\columnwidth]{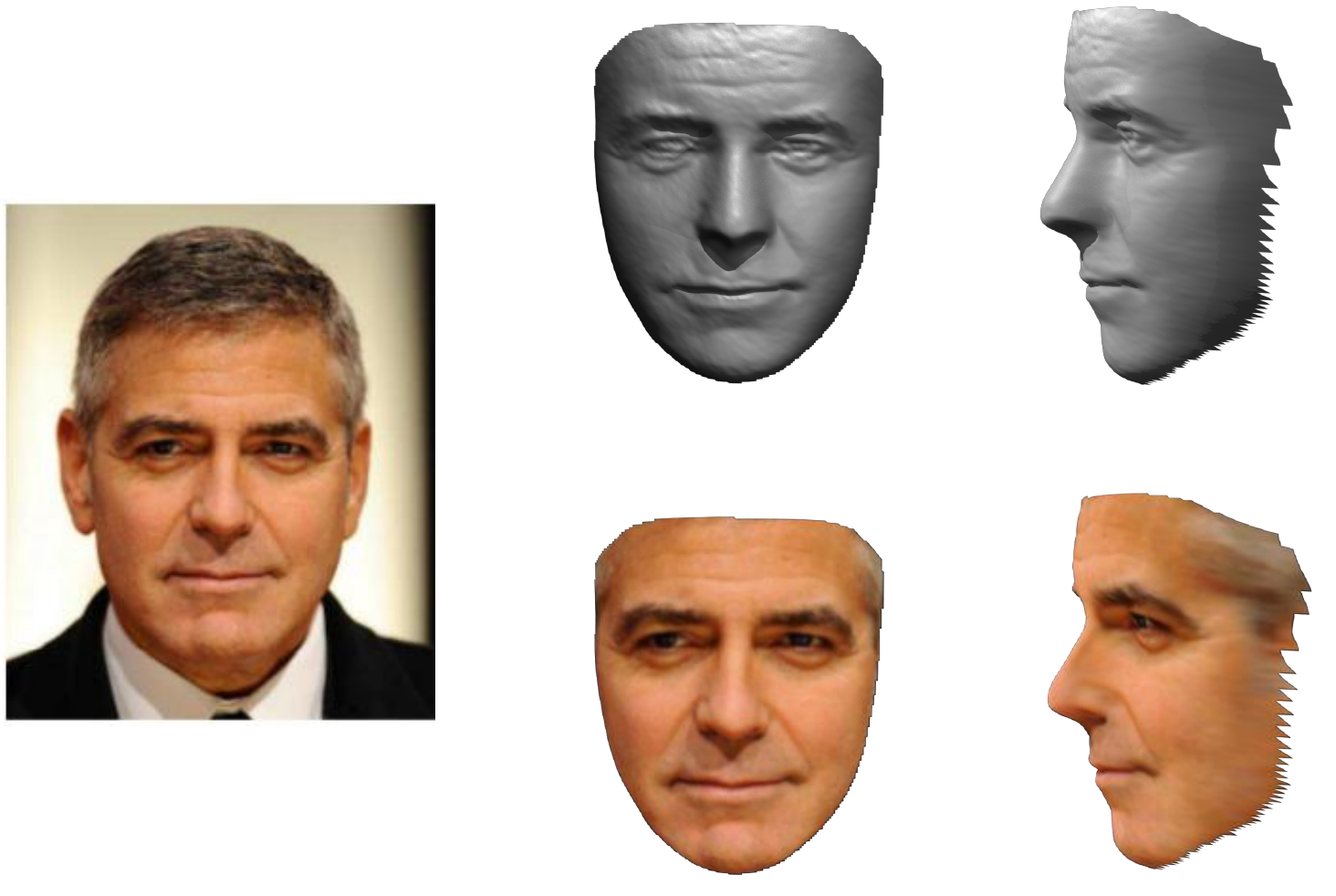}
	\caption{3D face reconstruction from a single image. Given an input image (left), we reconstruct a 3D face with fine geometric details (right, top row). The input image can be used as texture for rendering the reconstructed face (right, bottom row).}
	\label{fig:teaser}
\end{figure}

\begin{figure*}[t] 
	\centering
	\includegraphics[width=\textwidth]{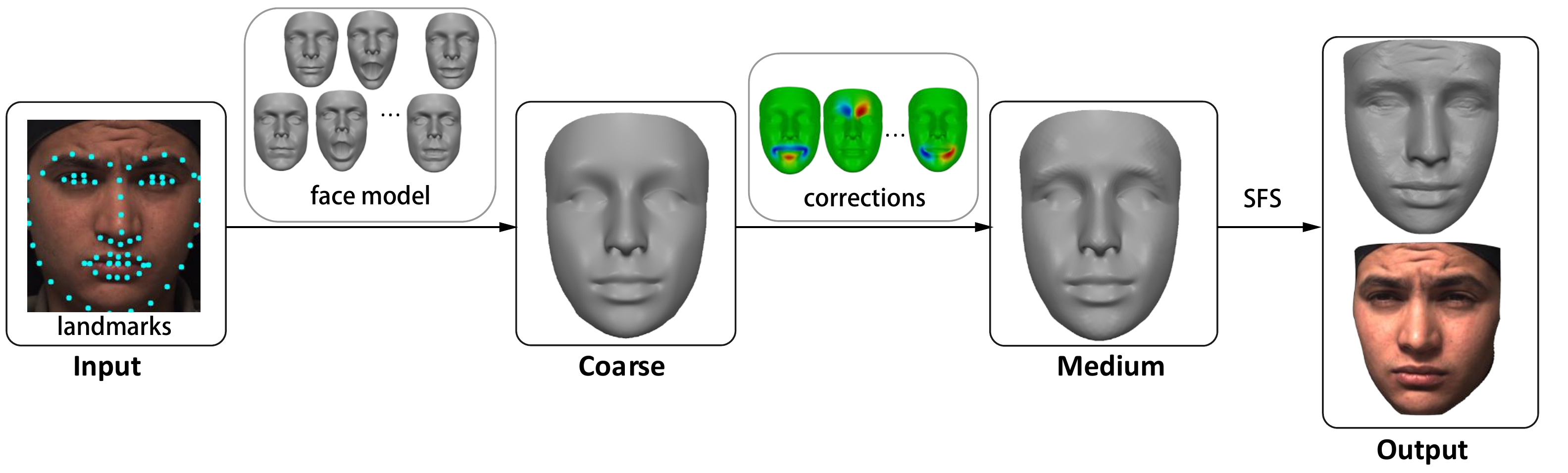}
	\caption{An overview of our coarse-to-fine face reconstruction approach.}
	\label{fig:pipeline}
\end{figure*}

In this paper, we propose a novel coarse-to-fine method to reconstruct a high-quality 3D face model from a single image. Our method consists of three steps:
\begin{itemize}
\item First, we compute a coarse estimation of the target 3D face, by fitting an example-based parametric face model to the input image. Our parametric model is derived from \facewarehouse{}~\cite{CaoWZTZ14} and \revise{the Basel Face Model (\bfm)}~\cite{Paysan2009}, two 3D face datasets with large variation in expression and identity respectively. The resulting mesh model captures the overall shape of the target face. 
\item Afterwards, we enhance the coarse face model by applying smooth deformation that captures medium-scale facial features; we also estimate the lighting and reflectance parameters from the enhanced face model. 
\item Finally, the illumination parameters and the enhanced face model are utilized to compute a height-field face surface according to the shading variation of the input image. This final model faithfully captures the fine geometric details of the target face (see Fig.~\ref{fig:teaser}).
\end{itemize}

Our method builds upon the strength of the existing approaches mentioned above: the example-based coarse face enables more reliable estimation of illumination parameters, and improves the robustness of the final SFS step; the SFS-based final face model provides detailed geometric features, which are often not available from example-based approaches. Our method outperforms existing example-based and SFS methods in terms of reconstruction accuracy as well as geometric detail recovery, as shown by extensive experimental results using publicly available datasets.

\section{Related Work}
\label{sec:Related}

\paraheading{Low-dimensional models}
Human faces have similar global characteristics, for example the location of main facial features such as eyes, nose and mouth. From a perception perspective, it has been shown that a face can be characterized using a limited number of parameters~\cite{Sirovich1987,Meytlis2007}. The low dimensionality of the face space allows for effective parametric face representations that are derived from a collection of sample faces, reducing the reconstruction problem into searching within the parameter space. A well-known example of such representations is the 3DMM proposed in~\cite{BlanzV99}, which has been used for various face processing tasks such as reconstruction\revise{~\cite{BlanzV99, romdhani2005estimating, keller20073d, bas2016fitting, huber2016multiresolution}}, recognition~\cite{BlanzV03,ZhuLYYL15}, face exchange in images~\cite{Blanz2004-Exchange}, and makeup suggestion~\cite{Scherbaum2011}. Low-dimensional representations have also been used for dynamic face  processing. To transfer facial performance between individuals in different videos, Vlasic et al.~\cite{VlasicBPP05} develop a multilinear face model representation that separately parameterizes different face attributes such as identity, expression, and viseme. In the computer graphics industry, facial animation is often achieved using linear models called blendshapes, where individual facial expressions are combined to create realistic facial movements~\cite{Lewis2014-Blendshape}. The simplicity and efficiency of blendshapes models enable real-time facial animation driven by facial performance captured from RGBD cameras~\cite{Weise2009,WeiseBLP11,BouazizWP13,li2013realtime,HsiehMYL15} and monocular videos~\cite{CaoWLZ13,CaoHZ14,CaoBZB15,Thies_2016_CVPR}. When using low-dimensional face representations derived from example face shapes, the example dataset has strong influence on the resulting face models. For instance, it would be difficult to reconstruct a facial expression that deviates significantly from the sample facial expressions. In the past, during the development of face recognition algorithms, various face databases have been collected and made publicly available~\cite{Gross2005}. Among them, \revise{\bfm{}} provides 3DMM representation for a large variety of facial identities. Recently, Cao et al.~\cite{CaoWZTZ14} introduced \facewarehouse{}, a 3D facial expression database that provides the facial geometry of 150 subjects, covering a wide range of ages and ethnic backgrounds. Our coarse face modeling method adopts a bilinear face model that encodes identity and expression attributes in a way similar to~\cite{VlasicBPP05}. We use \facewarehouse{} and \revise{\bfm{}} as the example dataset, due to the variety of facial expressions and identities that they provide respectively.

\paraheading{Shape-from-shading}
Shape-from-shading (SFS)~\cite{Zhang99shapefrom,Durou2008} is a computer vision technique that recovers 3D shapes from their shading variation in 2D images. Given the information about illumination, camera projection, and surface reflectance, SFS methods are able to recover fine geometric details that may not be available using low-dimensional models. On the other hand, SFS is an ill-posed problem with potentially ambiguous solutions~\cite{Prados2006}. Thus for face reconstruction, prior knowledge about facial geometry must be incorporated to achieve reliable results. For example, symmetry of human faces has been used by various authors to reduce the ambiguity of SFS results~\cite{Shimshoni2000,Zhao2000,Zhao2001}. Another approach is to solve the SFS problem within a human face space, using a low-dimensional face representation~\cite{Atick1996,Dovgard2004}. Other approaches improve the robustness of SFS by introducing an extra data source, such as a separate reference face~\cite{Kemelmacher-ShlizermanB11}, as well as coarse reconstructions using multiview stereo~\cite{WuWMT11,IchimBP15} or unconstrained photo collections~\cite{Kemelmacher-Shlizerman2011-InTheWild,RothTL15,roth2016adaptive}. We adopt a similar approach which builds an initial estimation of the face shape and augment it with fine geometric details using SFS. Our initial face estimation combines coarse reconstruction in a low-dimensional face space with refinement of medium-scale geometric features, providing a more accurate initial shape for subsequent SFS processing. 

\section{Overview}
\label{sec:Overview}

This section provides an overview of our coarse-to-fine approach to reconstructing a high-quality 3D face model from a single photograph. Fig.~\ref{fig:pipeline} illustrates the pipeline of our method.

To create a coarse face model (Sec.~\ref{sec:coarse}), we first build a bilinear model from \facewarehouse{} and \revise{\bfm{}} to describe a plausible space of 3D faces; the coarse face shape is generated from the bilinear model by aligning the projection of its 3D landmarks with the 2D landmarks detected on the input image, using a fitting energy that jointly optimizes the shape parameters (e.g., identity, expression) and camera parameters. To further capture person-specific features that are not available from the bilinear model, we enhance the coarse face using an additional deformation field that corresponds to medium-scale geometric features (Sec.~\ref{sec:medium}); the deformation field is jointly optimized with the lighting and albedo parameters, such that the shading of the enhanced model is close to the input image. Afterwards, the resulting medium face model is augmented with fine geometric details (Sec.~\ref{sec:fine}): the normal field from the medium face model is modified according to the input image gradients as well as the illumination parameters derived previously, and the modified normal field is integrated to achieve the final face shape. 

%Table~\ref{tab:symbols} lists the main symbols that are used in the following sections.
%\begin{table}[!htb]
%\caption{Symbols Used In This Paper.}
%\label{tab:symbols}
%\centering
%\scalebox{1.0}{
%\begin{tabular}{ll}
%\hline
%Symbol &Meaning\\
%\hline  
%$\bfF$ &The face shape which is a $3 \times 11510$ matrix\\        
%$\bfU$ & 2D landmarks assembled as a $2 \times 68$ matrix\\       
%$\Pi_{Q}$ &An orthographic projection matrix\\
%$\bfR$ &A $3 \times 3$ rotation matrix \\
%$\bft$ &A $2 \times 1$ translation vector \\
%$\bfw_{c}$ &The camera parameters including $\Pi_{Q}$, $\bfR$, $\bft$\\
%$\bfw_{id}$ &A column vector of the identity weights \\
%$\bfw_{exp}$ &A column vector of the expression weights \\
%$\bfI_{i,j}$ &The gray-scale value at the pixel$\{i, j\}$ \\
%$\bfs_{i,j}$ &The illumination at the pixel $\{i, j\}$ \\
%$\bfn_{i,j}$, $\bfN$ &The normal at the pixel$\{i, j\}$, $\bfN$ stack all the $\bfn_{i,j}$ \\
%$\bfr_{i,j}$, $\bfr$ &The skin reflectance at the pixel$\{i, j\}$, $\bfr$ stack all the $\bfr_{i,j}$ \\
%$\bfz_{i,j}$, $\bfz$ &The depth value at the pixel$\{i, j\}$, $\bfz$ stack all the $\bfz_{i,j}$ \\
%$\bfG$ &The gradient matrix \\
%$\bfL$ &The Laplacian matrix \\
%\hline
%\end{tabular}}
%\end{table}

\section{Coarse Face Modeling}
\label{sec:coarse}
\paraheading{Preprocessing}
\revise{The \facewarehouse{} dataset contains head meshes of 150 individuals, each with 47 expressions. All expressions are represented as meshes with the same connectivity, each consisting of 11510 vertices. The \bfm{} dataset contains 200 face meshes, and each mesh consists of 53490 vertices. In order to combine the two datasets, \revise{we first mask the face region on the head mesh from \facewarehouse{} to extract a face mesh, and fill the holes in the regions of eyes and mouth, to obtain a simply connected face mesh consisting of 5334 vertices.} Afterwards, we randomly sample the parameter space for \bfm{} to generate 150 neutral face models, and deform the average face model from \facewarehouse{} to fit these models via nonrigid registration~\cite{sd2004}. Then we transfer the other 46 expressions of the \facewarehouse{} average face model to each of the 150 deformed face models based on the method in~\cite{sd2004}. In this way, we construct a new dataset containing 300 individuals (150 from \bfm{} and 150 from \facewarehouse{}), each with 47 expressions. We perform Procrustes alignment for all the face meshes in the dataset. Moreover, \bfm{} provides 199 principal components to span the surface albedo space, but these principal albedo components cannot be used for our new dataset directly due to different mesh connectivity. Thus we transfer their albedo information to the new mesh representation using the correspondence identified in the nonrigid registration, to construct 199 principal albedo components for our dataset. These principal components will be used in Sec~\ref{sec:medium}.}

\paraheading{Bilinear face model}
Following~\cite{VlasicBPP05}, we collect the vertex coordinates of \revise{all face meshes} into a third-order data tensor, and perform $2$-mode SVD reduction along the identity mode and the expression mode, to derive a bilinear face model that approximates the original data set. In detail, the bilinear face model is represented as a mesh with the same connectivity as those from the data set, and its vertex coordinates $\bfF \in \mathbb{R}^{3 \times N_v}$ are computed as
\begin{equation}\label{eq:dataset approximate}
\bfF=\bfC_{\textrm{r}} \times_{2} \wid^{T}\times_{3} \wexp^{T},
\end{equation}
where \revise{$N_v$ is the number of vertices}, $\bfC_{\textrm{r}}$ is the reduced core tensor computed from the SVD reduction, and $\wid{} \in \mathbb{R}^{100}, \wexp{} \in \mathbb{R}^{47}$ are column vectors for the identity weights and expression weights which control the face shape. Note that here we only reduce the dimension along the identity mode, in order to maintain the variety of facial expressions in the bilinear model. For more details on multilinear algebra, the reader is referred to~\cite{de1997signal}.

\begin{figure}[t] 
   \centering
      \includegraphics[width=0.85\columnwidth]{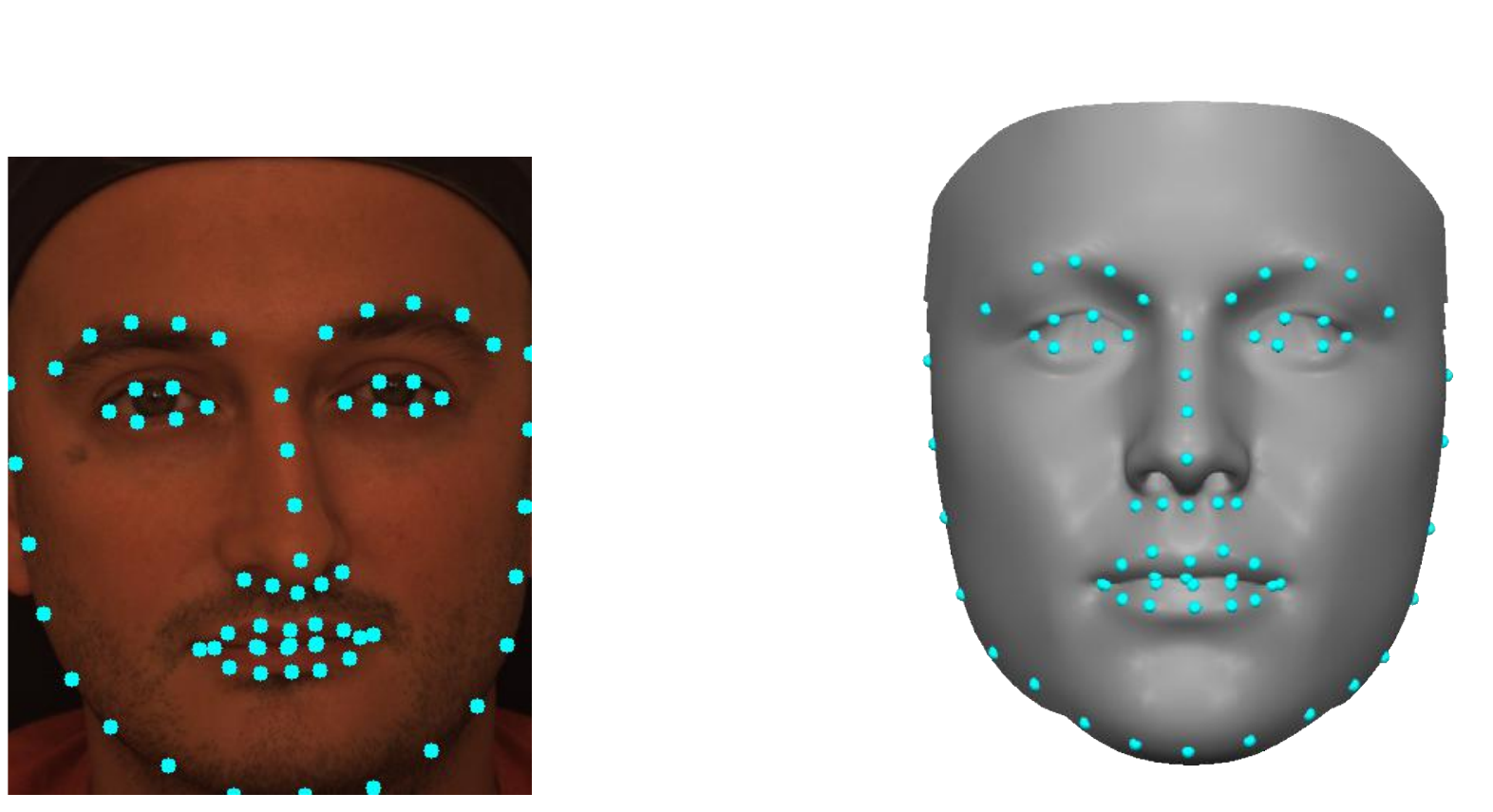}
   \caption{Our coarse face reconstruction is based on aligning the projection of labeled 3D face landmarks (right) with 2D landmarks detected on the input image (left).}
  \label{fig:landmarks}
 \end{figure}

To construct a coarse face, we align 3D landmarks on the bilinear face model with corresponding 2D landmarks from the input image. First, we preprocess the bilinear face mesh to manually label 68 landmark vertices. Given an input image, we detect the face as well as its corresponding 68 landmarks using the method in~\cite{ChenRWCS14} (see Fig.~\ref{fig:landmarks} for an example). Assuming that the camera model is a weak perspective projection along the $Z$ direction, we can write the projection matrix as
$\proj=
\left[
\begin{matrix}
\alpha & 0 & 0 \\
0 & \alpha & 0 \\
\end{matrix}
\right]
$.
Then we can formulate the following fitting energy to align the projection of landmark vertices with the detected 2D landmarks
\begin{eqnarray}
\efit{} &=& \sum_{k=1}^{68} \left\| \proj{} \bfR \bfF_{v_k} + \bft - \bfU_{k} \right\|_2^2 \nonumber\\ 
&& + \: \weightfit{1}  \sum_{i=1}^{100} \left( \frac{\widc{i}}{\didc{i}}  \right)^2 + \weightfit{2} \sum_{j=1}^{47} \left( \frac{\wexpc{j}}{\dexpc{j}}  \right)^2. 
\label{eq:2D3DFitting}
\end{eqnarray}
Here $\bfF_{v_k} \in \mathbb{R}^{3}$ and $\bfU_{k} \in \mathbb{R}^{2}$ are the coordinates of the $k$-th 3D landmark vertex and the corresponding image landmark, respectively; translation vector $\bft \in \mathbb{R}^2$ and rotation matrix $\bfR \in \mathbb{R}^{3 \times 3}$ determine the position and pose of the face mesh with respect to the camera; $\widc{i}$ and $\wexpc{j}$ are components of weight vectors $\wid{}$ and $\wexp{}$, while $\didc{i}$ and $\dexpc{j}$ are the corresponding singular values obtained from the $2$-mode SVD reduction; $\weightfit{1}$ and $\weightfit{2}$ are positive weights. As in~\cite{BlanzV99}, the last two terms ensure parameters $\widc{i}$ and $\wexpc{j}$ have a reasonable range of variation. 
This fitting energy is minimized with respect to the shape parameters $\wid{}, \wexp{}$ and the camera parameters $\proj,\bfR,\bft$ via coordinate descent. 
%First, the shape parameters can be initialized as
%\begin{equation}\label{eq:initialization of shape parameters}
%\bfw_{id}=\frac{1}{50}\sum_{m=1}^{50}\bfw_{id}^{m}, \:\:\: \bfw_{exp}=\frac{1}{47}\sum_{n=1}^{47}\bfw_{exp}^{n},
%\end{equation}
%where $\bfw_{id}^{m}$ and $\bfw_{exp}^{n}$ mean the $m$-th identity weight and the $n$-th expression weight in the database respectively. 
First we fix the shape parameters and reduce the optimization problem to
\begin{equation}\label{eq:solving the camera parameters}
\min_{\proj,\bfR,\bft} ~~\sum_{k=1}^{68} \left\| \proj{} \bfR \bfF_{v_k} + \bft - \bfU_{k} \right\|_2^2,
\end{equation}
which is solved using the pose normalization method from ~\cite{Kemelmacher-Shlizerman2011-InTheWild}. 
Next we fix the camera and expression parameters, which turns the optimization into
\begin{equation}\label{eq:solving the shape parameters}
\min_{\wid{}} ~~ \sum_{k=1}^{68} \left\| \proj{} \bfR \bfF_{v_k} + \bft - \bfU_{k} \right\|_2^2 + \weightfit{1}  \sum_{i=1}^{100} \left( \frac{\widc{i}}{\didc{i}}  \right)^2.
\end{equation}
This is a linear least-squares problem and can be easily solved by solving a linear system. Finally, we fix the camera and identity parameters, and optimize the expression parameters in the same way as \revise{Eq.~\eqref{eq:solving the shape parameters}}. These steps are iteratively executed until convergence. In our experiments, four iterations are sufficient for convergence to a good result.
 
\begin{figure}[t] 
   \centering
      \includegraphics[width=0.75\columnwidth]{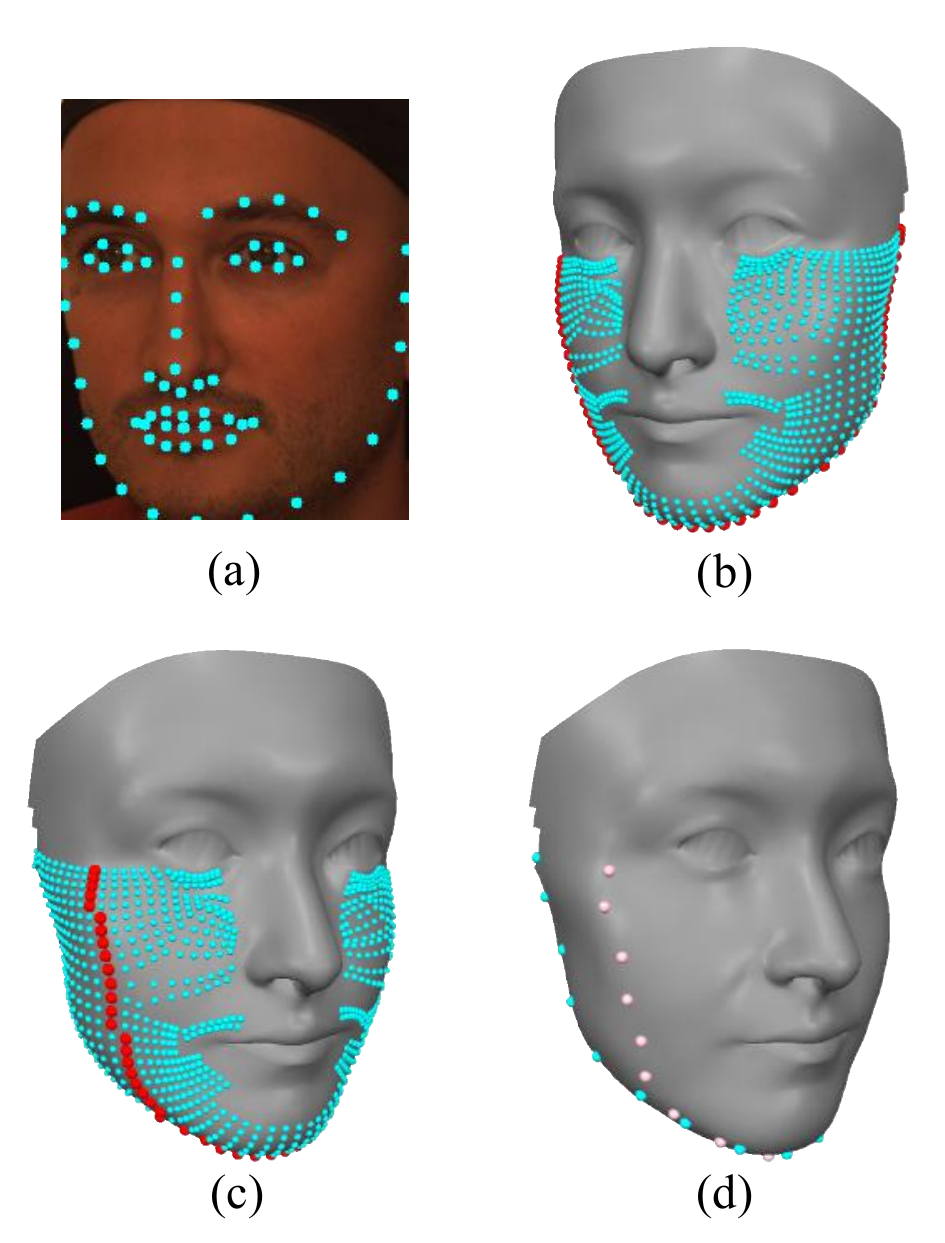}
   \caption{\revise{For a non-frontal face images (a), the labeled 3D face silhouette landmarks (shown in cyan in (d)) need to be updated for better correspondence with the detected 2D silhouette landmarks. We construct a set of horizontal lines connecting the mesh vertices (shown in cyan in (b) and (c)), and select among them a set of vertices representing the updated silhouette according to the current view direction (shown in red in (b) and (c)). The new 3D silhouette landmarks (shown in pink in (d)) are selected within the updated silhouette.}}
  \label{fig:updated_landmarks}
 \end{figure} 
\paraheading{Landmark vertex update}
The landmark vertices on the face mesh are labeled based on the frontal pose. For non-frontal face images, the detected 2D landmarks along the face silhouette may not correspond well with the landmark vertices (see Fig.~\ref{fig:updated_landmarks}(a) for an example). Thus after each camera parameter optimization step, we update the silhouette landmark vertices according to the rotation matrix $\bfR$, while keeping the internal landmark vertices (e.g., those around the eyes, the nose, and the mouth) unchanged. Similar to~\cite{CaoHZ14}, we preprocess the original face mesh to derive a dense set of horizontal lines that connect mesh vertices and cover the potential silhouette region from a rotated view (see Fig.~\ref{fig:updated_landmarks}(b) and~\ref{fig:updated_landmarks}(c)). Given a rotation matrix $\bfR$, we select from each horizontal line a vertex that lies on the silhouette, and project it onto the image plane according to the camera parameters $\proj,\bfR,\bft$. These projected vertices provide an estimate of the silhouette for the projected face mesh. Then for each 2D silhouette landmark, its corresponding landmark vertex is updated to the silhouette vertex whose projection is closest to it (see Fig.~\ref{fig:updated_landmarks}(d)).

To determine the silhouette vertex on a horizontal line, we select the vertex whose normal encloses the largest angle with the view direction. Since the face mesh is approximately spherical with its center close to the origin, we approximate the unit normal of a vertex on the rotated face mesh as $\frac{\bfR \bfv}{\|\bfR \bfv\|_2}$, where $\bfv$ is the original vertex coordinates. Then the silhouette vertex is the one with the smallest value of $\left|  \projdirection \cdot \frac{\bfR \bfv}{\|\bfR \bfv\|_2} \right|$ within the horizontal line, where $\projdirection = [0, 0, 1]^T$ is the view direction. 
\begin{figure}[htbp] 
   \centering
      \includegraphics[width=0.90\columnwidth]{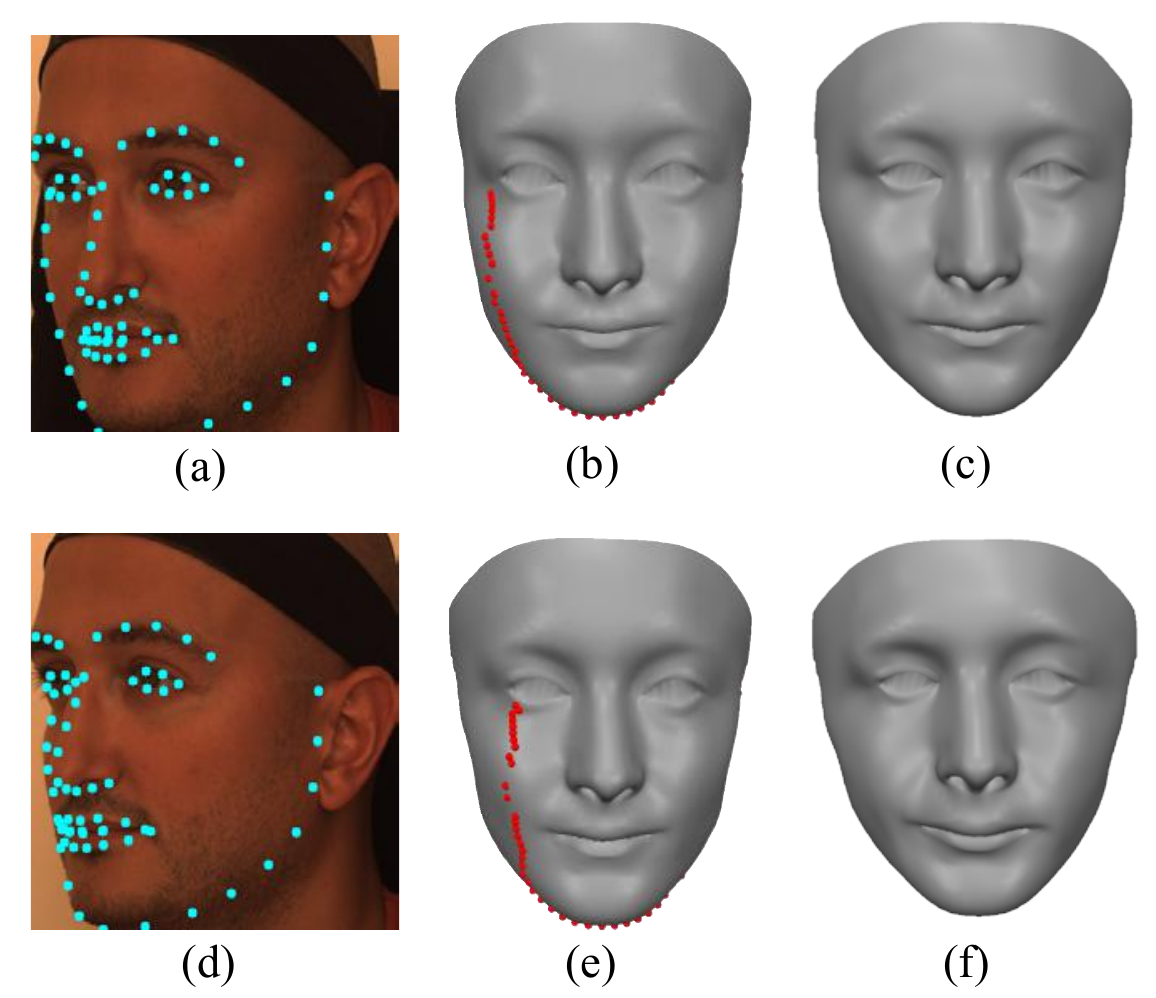}
   \caption{\revise{Silhouette update improves accuracy of the coarse face model. Each row shows an input image ((a) and (d)), the corresponding coarse face model with silhouette update ((b) and (e)), and the one without silhouette update ((c) and (f)). The updated silhouette is shown in red. The top row shows an example with $+30^{\circ}$ yaw, and the bottom row with $+45^{\circ}$ yaw.}}
  \label{fig:differentpose}
 \end{figure}
 
\revise{The silhouette update improves the accuracy of the coarse face model for non-frontal images, as shown in Fig.~\ref{fig:differentpose} for two examples with $+30^{\circ}$ and $+45^{\circ}$ yaws: without the silhouette update, the resulting model will become wider due to erroneous correspondence with between the detected landmarks and the silhouette landmarks. When the yaw becomes larger, the detected 2D landmarks become less reliable, and the coarse face model becomes less accurate even with silhouette update. Our approach does not work well for images with very large poses (beyond $60^{\circ}$ yaw) unless the invisible landmarks can be accurately detected. On the other hand, our pipeline can be combined with large-pose landmark detection algorithms to produce good results for such images. Some examples are shown in Fig.~\ref{fig:bigpose}.}

\section{Medium Face Modeling}
\label{sec:medium}

\begin{figure*}[t] 
   \centering
      \includegraphics[width=\textwidth]{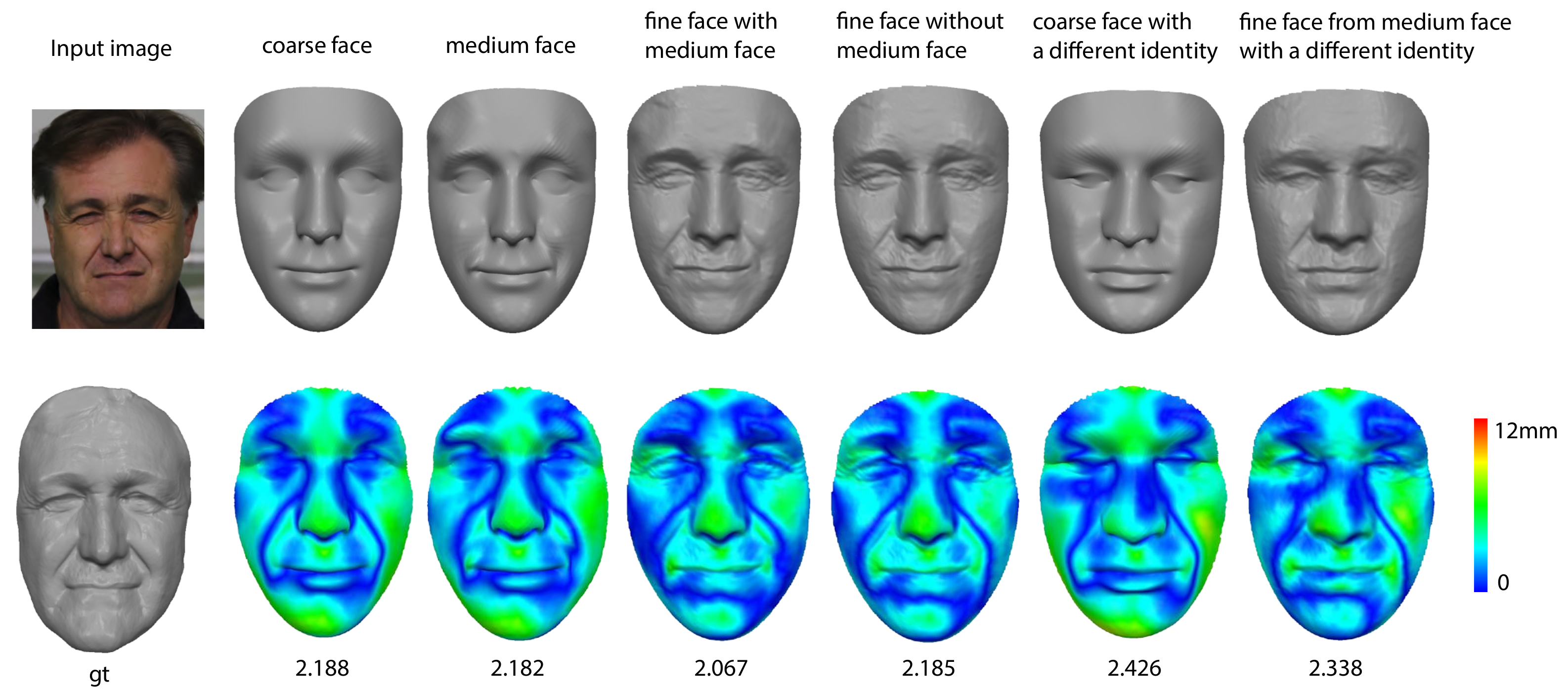}
   \caption{Quantitative results on the dataset~\cite{ValgaertsWBST12}. The input image and its ground truth shape are shown in the first column. In the other columns, we show different face reconstructions and their corresponding error maps (according to Eq.~\eqref{eq:ErrorMap}): the coarse face model, the medium face model, the fine reconstruction with and without medium face modeling, the coarse model with modified identity parameters, and the fine reconstruction with medium face modeling from the modified coarse face. In the bottom, we show the reconstruction error values.}
  \label{fig:enhance}
 \end{figure*}

Although the coarse face model provides a good estimate of the overall shape, it may not capture some person-specific geometric details due to limited variation of the constructed data set (see Fig.~\ref{fig:enhance}). Thus we enhance the coarse face using smooth deformation that correspond to medium-scale geometric features, to improve the consistency between its shading and the input image. During this process we also estimate the lighting and the albedo. The enhanced face model and the lighting/albedo information will provide the prior knowledge required by the SFS reconstruction in the next section. In this paper, we convert color input images into grayscale ones for simplicity and efficiency. However, it is not difficult to extend the formulation to directly process color images.
 
\paraheading{Lighting and albedo estimation}
To compute shading for our face mesh, we need the information about lighting and surface reflectance. Assuming Lambertian reflectance, we can approximate the grayscale level $s_{i,j}$ at a pixel $(i,j)$ using second-order spherical harmonics~\cite{frolova2004accuracy}:
\begin{equation}
\label{eq:sphericalharmoniclighting}
s_{i,j} = r_{i,j} \cdot \max(\sphcoef^T \bfH(\bfn_{i,j}), 0).
\end{equation}
Here $r_{i,j}$ is the albedo at the pixel; $\bfn_{i,j}$ is the corresponding mesh normal, computed via
\begin{equation}
	\bfn_{i,j} = \frac{(\bfv_2^{i,j} - \bfv_1^{i,j}) \times (\bfv_3^{i,j} - \bfv_1^{i,j})}{\|(\bfv_2^{i,j} - \bfv_1^{i,j}) \times (\bfv_3^{i,j} - \bfv_1^{i,j})\|_2},
	\label{eq:meshnormal}
\end{equation}
where $\bfv_1^{i,j}, \bfv_2^{i,j}, \bfv_3^{i,j}$ are the vertex coordinates for the mesh triangle that corresponds to pixel $(i,j)$;
$\bfH$ is a vector of second-order spherical harmonics
\begin{equation}\label{eq:the second spherical harmonics}
 \bfH(\bfn) = [1, n_x, n_y, n_z, n_xn_y, n_xn_z, n_yn_z, n_x^2-n_y^2, 3n_z^2-1]^T, 
\end{equation}
and $\bm{\xi}$ is a vector of harmonics coefficients. For more robust estimation, we follow~\cite{BlanzV99} and parametrize the surface reflectance using a \emph{Principal Component Analysis} (PCA) model:
\begin{equation}
\label{eq:albedo pcd model}
r_{i,j} = \left(\bm{\Phi}_{0} + \sum_{l=1}^{N_r} w_{r}^{l}\bm{\Phi}_{l} \right) \cdot \mathbf{c}_{i,j},
\end{equation}
where $[c_{i,j}^{1}, c_{i,j}^{2}, c_{i,j}^{3}] \in \mathbb{R}^{3}$ is the barycentric coordinate of the triangle corresponding to $r_{i,j}$, $[\bm{\Phi}_0, \bm{\Phi}_1,...,\bm{\Phi}_{N_r}] \in \mathbb{R}^{N_v \times (N_r+1)}$ is a basis of vertex albedos with $N_v$ being the number of vertices of the face mesh, $\bfw_r=(w_{r}^{1},...,w_{r}^{N_r}) \in \mathbb{R}^{N_r}$ is a vector for the albedo weights; $\mathbf{c}_{i,j} \in \mathbb{R}^{N_v}$ is a vector whose components for the three vertices of the triangle that contains pixel $(i,j)$ are equal to the barycentric coordinates of the pixel within the triangle, and the components for other vertices are zero. \revise{Among the 199 principal albedo components derived from \bfm{}, we choose $N_r$ principal components with the largest variance as $\bm{\Phi}_1,...,\bm{\Phi}_{N_r}$. We set $N_r=100$ in our experiments.} The lighting and albedo are then estimated by solving an optimization problem
\begin{eqnarray}
\min_{\bfr, \bm{\xi}, \bfd} && \sum_{i,j} \left( r_{i,j}\sphcoef^T \bfH(\bfn_{i,j}) - I_{i,j} \right)^2 +\weightalbedo{1}\sum_{l = 1}^{N_r} \left\| \frac{w_{r}^{l}}{\drefc{l}} \right\|_2^2,
\label{eq:AlbedoLightingOptimization}
\end{eqnarray}
where vectors $\bfr, \bfd$ collect the values $\{r_{i,j}\}, \{d_{i,j}\}$, respectively; $I_{i,j}$ denotes the grayscale value at pixel $(i,j)$ of the input image; $\{\drefc{l}\}$ are the standard deviations corresponding to the principal directions; $\weightalbedo{1}$ is a user-specified positive weight. To optimize this problem, we first set $\bfw_r$ to zero and optimize the harmonics coefficients $\bm{\xi}$. Then we optimize the reflectance weights $\bfw_r$ while fixing $\bm{\xi}$. Both sub-problems reduce to solving a linear system. This process is iterated three times in our experiment.

\begin{figure}[t] 
   \centering
      \includegraphics[width=\columnwidth]{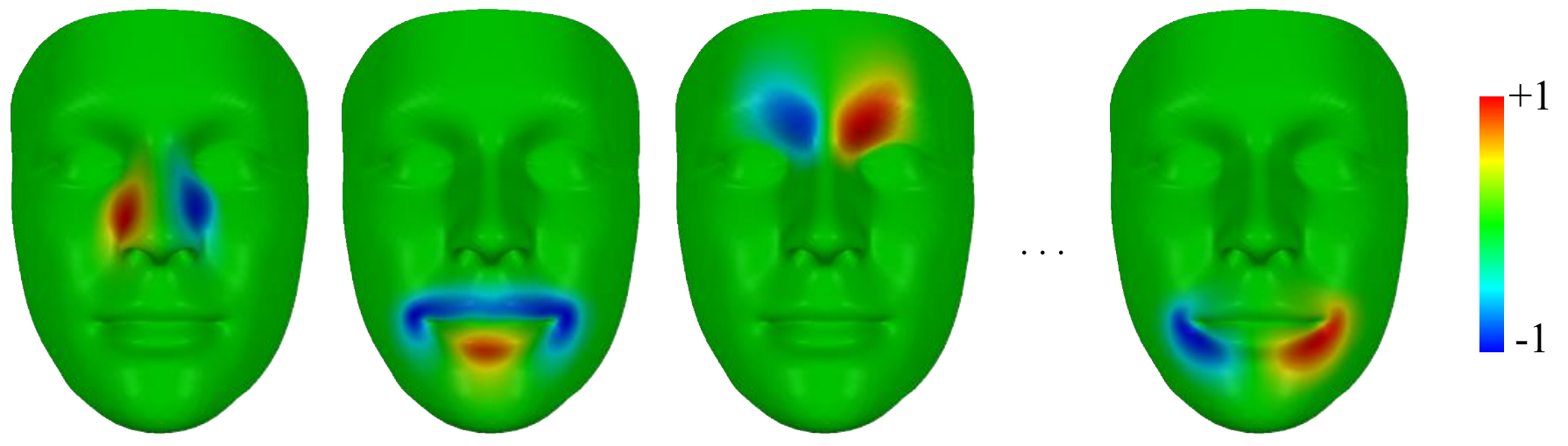}
   \caption{Some Laplacian eigenfunctions of local regions on the face mesh (displayed via color coding).}
  \label{fig:corrections}
 \end{figure}

\paraheading{Facial detail enhancement}
With an estimate of lighting and albedo, we can now enhance the coarse face mesh to reduce the discrepancy between the mesh shading and the input image. We apply a smooth 3D deformation field to the $N_v$ vertices of the frontal face mesh to minimize the following discrepancy measure with respect to the vertex displacements $ \bfD \in \mathbb{R}^{3 \times N_v}$:
\begin{equation}
	\eshade{}(\bfD) = \sum_{i,j} \left( r_{i,j} \max(\sphcoef^T \bfH(\widetilde{\bfn}_{i,j}), 0)  - I_{i,j}\right)^2,
	\label{eq:deformationproblem}
\end{equation}
where $\{\widetilde{\bfn}_{i,j}\}$ are the new mesh face normals. \revise{Specifically, since our final goal is to recover a depth field defined on the facial pixels in the given image, we sum over the pixels in Eq.~\eqref{eq:deformationproblem}. The correspondence between pixels and triangles are computed by the Z-buffer method~\cite{strasser1974schnelle}.} However, this nonlinear least-squares problem can be very time-consuming to solve, due to the high resolution of the mesh. Therefore, we construct a low-dimensional subspace of smooth mesh deformations and solve the optimization problem within this subspace, which significantly reduces the number of variables. Specifically, if we measure the smoothness of a deformation field using the norm of its graph Laplacian with respect to the mesh, then the Laplacian eigenfunctions associated with small eigenvalues span a subspace of smooth deformations. Indeed, it is well known in 3D geometry processing that the Laplacian eigenvalues can be seen as the frequencies for the eigenfunctions, which indicate how rapidly each eigenfunction oscillates across the surface~\cite{Zhang2010-Spectral}. Thus by restricting the deformation to the subspace with small eigenvalues, we inhibit the enhancement of fine-scale geometric features, leaving them to the SFS reconstruction step in Sec~\ref{sec:fine}. Since most facial variations are local, we select some local regions on the mesh, and perform Laplacian eigenanalysis on each region separately (see Fig.~\ref{fig:corrections}). The selected eigenfunctions are then combined to span a space of facial variations.
Specifically, for the $i$-th selected region, we preprocess the frontal face mesh to construct its graph Laplacian matrix $ \graphlap \in \mathbb{R}^{N_v \times N_v}$ based on mesh connectivity, and add a large positive value to the $j$-th diagonal element if vertex $j$ is outside the selected region. Then we perform eigendecomposition to obtain $k + 1$ eigenvectors $\bfe_0^i, \bfe_1^i, \ldots, \bfe_k^i$ corresponding to the smallest eigenvalues $\lambda_0^i \leq \lambda_1^i \leq \ldots \leq \lambda_k^i$. Among them, $\bfe_0^i$ has a constant value inside the selected region, representing a translation of the whole region~\cite{Zhang2010-Spectral}. Since it does not represent variation within the region, we discard $\bfe_0^i$ to get $k$ eigenvectors $\bfE^i=[\bfe_1^i, \ldots, \bfe_k^i]$. Combing all the eigenvectors to span the $x$-, $y$-, and $z$-coordinates of the vertex displacement vectors, we represent the deformation field as
\begin{equation}
	\bfD = (\bfE \bm{\eta})^T,
	\label{eq:subspacedeformation}
\end{equation}
where $\bfE = [ \bfE^1, \ldots, \bfE^{N_e} ] \in \mathbb{R}^{N_v \times (k \cdot N_e)}$ stacks the basis vectors, and 
$\bm{\eta} = [ \lambda_1^1, \ldots, \lambda_k^1, \ldots, \lambda_1^{N_e}, \ldots, \lambda_k^{N_e} ]^T \in \mathbb{R}^{(k \cdot N_e) \times 3}$ collects their linear combination coefficients. Then the deformation is determined by solving the following optimization problem about $\bm{\eta}$:
\begin{equation}
	\min_{\bm{\eta}} ~~ \eshade{} (\bfD) + \weightsubspace \sum_{i=1}^{N_e} \sum_{j=1}^k \left\| \frac{\bm{\eta}_j^i}{\lambda_j^i} \right\|_2^2.
	\label{eq:ReducedDeformationOptimization}
\end{equation}

Here the second term prevents large deformations, with more penalty on basis vectors of lower frequencies; $\weightsubspace$ is a user-specified weight. Our formulation is designed to induce more enhancement for finer geometric features, since the coarse face already provides a good estimate of the overall shape. \revise{In our experiments, we set $k=5$ and $N_e=9$, which means we select nine local regions and the first five eigenfunctions of the corresponding Laplacian matrix for each region. These local regions are manually selected in a heuristic way. More specifically, given the mean face shape, we first compute the vertex displacements from its neutral expression to each of the other 46 expressions, and manually select nine regions with the largest variation as the local regions.}

As the number of variables are significantly reduced in \eqref{eq:ReducedDeformationOptimization}, this nonlinear least-squares problem can be solved efficiently using the Levenberg-Marquardt algorithm~\cite{Madsen2004}. We then apply the optimized deformation field to the frontal face mesh, and update the correspondence between image pixels and mesh triangles. With the new correspondences, we solve the optimization problems \eqref{eq:AlbedoLightingOptimization} and \eqref{eq:ReducedDeformationOptimization} again to further improve the lighting/albedo estimate and the face model. This process is iterated twice in our experiments.

\begin{figure}[t] 
   \centering
      \includegraphics[width=\columnwidth]{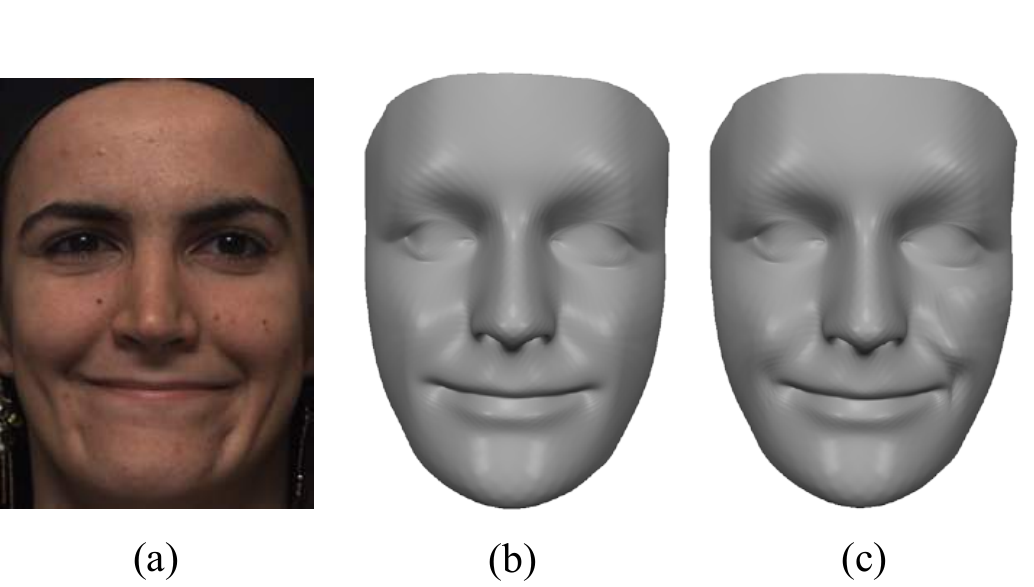}
   \caption{\revise{An input image with smile expression (a), and its coarse (b) and medium (c) face models. The use of Laplacian eigenvectors improves the accuracy of features around the laugh lines, but cannot reconstruct the dimples as the eigenvectors provide limited cover of their frequency band.}}
  \label{fig:basisfunction}
 \end{figure}

\revise{Medium face modeling can improve the accuracy of medium-scale facial features such as those around the laugh lines, as shown in Figs.~\ref{fig:enhance} and Figs.~\ref{fig:basisfunction}. Fig.~\ref{fig:enhance} compares the fine face reconstruction results with and without medium face modeling. We can see that the use of medium face leads to more accurate results numerically and visually. Indeed, eigendeomposition of the Laplacian matrix corresponds to Fourier analysis of geometric signals defined on the mesh surface~\cite{Zhang2010-Spectral}, thus our use of basisvectors is similar to approximating the displacement from the coarse face to the ground truth shape in each local region using its Fourier components of lowest frequencies, which is a classical signal processing technique. On the other hand, our approach cannot reconstruct facial features whose frequency bands have limited overlap with those corresponding to the chosen basisvectors. One example is shown in Fig.~\ref{fig:basisfunction}, where the dimples cannot be reconstructed. Finally, as the medium face modeling is applied on local regions, it cannot reduce reconstruction errors of global scales. As an example, in Fig.~\ref{fig:enhance} we alter the identity parameters to generate a different coarse face model, and apply medium and fine face modeling. We can see that although medium and fine face modeling help to introduce more details, they cannot change the overall face shape.}
	
%	Fig.~\ref{fig:enhance} shows the reconstructed face model from each algorithm stages. We can observe that the deformation fields help to improve the reconstruction accuracy both numerically and visually. The laugh lines at the corners of the mouth of the medium face are more obvious than the coarse face in Fig.~\ref{fig:enhance}. The reconstruction improvement by the deformation fields can be also observed in Fig.~\ref{fig:basisfunction}. The coarse reconstruction can capture the global structure of face shape very well, but it is not good enough to describe common local shape variations (e.g., laugh lines), and cannot describe uncommon local shape variations (e.g., dimples) at all (see Fig.~\ref{fig:basisfunction}(a) and Fig.~\ref{fig:basisfunction}(b)). Our introduced deformation fields can help the coarse faces to have a more obvious description of common local variations (see Fig.~\ref{fig:basisfunction}(c)). Since our deformation fields contain only low, medium variations, we cannot recover uncommon local shape variations in the medium face modeling. And for face images with small shape variations (see Fig.~\ref{fig:basisfunction}(d)), the coarse face models could be good enough (see Fig.~\ref{fig:basisfunction}(e)), thus our deformation fields have a little help (see Fig.~\ref{fig:basisfunction}(f)).}

\section{Fine Face Modeling}
\label{sec:fine}

As the final step in our pipeline, we reconstruct a face model with fine geometric details, represented as a height field surface over the face region $\faceregion$ of the input image. Using the medium face model and the lighting/albedo information computed in Sec.~\ref{sec:medium}, we first compute a refined normal map over $\faceregion$, to capture the details from the input image. This normal map is then integrated to recover a height field surface for the final face shape.

\paraheading{Overall approach}
Specifically, the normal map is defined using a unit vector $\normalmap{i,j} \in \mathbb{R}^3$ for each pixel $(i,j) \in \faceregion$. Noting that each face pixel corresponds to a normal vector facing towards the camera~\cite{Kemelmacher-ShlizermanB11}, we represent $\normalmap{i,j}$ using two variables $p_{i,j}, q_{i,j}$ as
\begin{equation}
	\normalmap{i,j} = \frac{(p_{i,j},q_{i,j},-1)}{\sqrt{p_{i,j}^2+q_{i,j}^2+1}}.
	\label{eq:pixelnormal}
\end{equation}
The values $\{p_{i,j}\}$, $\{q_{i,j}\}$ are computed by solving an optimization problem that will be explained later. The final height-field face model, represented using a depth value $z_{i,j}$ per pixel, is then determined so that the height field normals are as close as possible to the normal map. We note that the height field normal $\heightfieldnormal{i,j}$ at pixel $(i,j)$ can be computed using three points $\mathbf{h}_{i,j} = (i, j, z_{i,j})$, $\mathbf{h}_{i,j+1} = (i, j+1, z_{i, j+1})$, $\mathbf{h}_{i+1,j} = (i+1, j, z_{i+1,j})$ on the height field surface via
\begin{eqnarray}
	\heightfieldnormal{i,j} &=& \frac{ (\mathbf{h}_{i,j+1}  - \mathbf{h}_{i,j}) \times (\mathbf{h}_{i+1,j}  - \mathbf{h}_{i,j}) }
			{\|(\mathbf{h}_{i,j+1}  - \mathbf{h}_{i,j}) \times (\mathbf{h}_{i+1,j}  - \mathbf{h}_{i,j})\|_2} \nonumber\\
			&=& \frac{\left(z_{i+1,j}-z_{i,j}, z_{i,j+1}-z_{i,j}, -1\right)}{\sqrt{(z_{i+1,j}-z_{i,j})^2+(z_{i,j+1}-z_{i,j})^2+1}}.
\end{eqnarray}
Comparing this with Eq.~\eqref{eq:pixelnormal} shows that for the height field normal to be consistent with the normal map, we should have
\begin{equation}
	z_{i+1,j}-z_{i,j} = p_{i,j}, \quad z_{i,j+1}-z_{i,j} = q_{i,j}
	\label{eq:HeightFieldConditions}
\end{equation}
for every pixel. As these conditions only determine $\{z_{i,j}\}$ up to an additional constant, we compute $\{z_{i,j}\}$ as the minimum-norm solution to a linear least-squares problem
\begin{equation}
	\min_{\{z_{i,j}\}} ~~ \sum_{(i,j)} (z_{i+1,j}-z_{i,j} - p_{i,j})^2 + (z_{i,j+1}-z_{i,j} - q_{i,j})^2.
	\label{eq:DepthReconstruction}
\end{equation}

\begin{figure}[t] 
   \centering
      \includegraphics[width=\columnwidth]{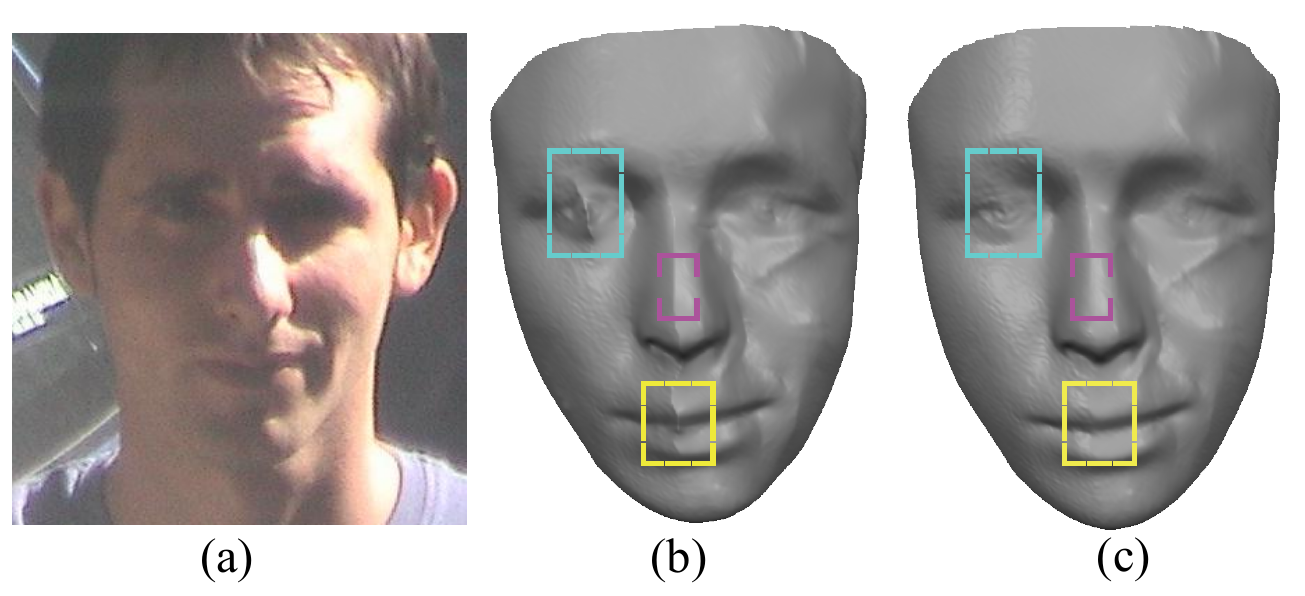}
   \caption{\revise{An input image with cast shadow and noise (a), and its reconstruction results by minimizing the intensity difference (b) and the gradient difference (c), respectively. Compared with intensity difference minimization, the formulation with gradient difference produces a smoother result and reduces unnatural sharp creases at the eye, the nose, and the mouth (highlighted with rectangles).}}
  \label{fig:shadow_cast}
 \end{figure}

\paraheading{Normal map optimization}
For high-quality results, we enforce certain desirable properties of the computed normal map $\normalmap{i,j}$ by minimizing an energy that corresponds to these properties. First of all, the normal map should capture fine-scale details from the input image. Using the lighting and albedo parameters obtained during the computation of the medium face, we can evaluate the pixel intensity values from the normal map according to Eq.~\eqref{eq:sphericalharmoniclighting}, and require them to be close to the input image. However, such direct approach can suffer from the inaccuracy of spherical harmonics in complex lighting conditions such as cast shadows, which can lead to unsatisfactory results. Instead, we aim at minimizing the difference in intensity gradients, between the input image and the shading from the normal map. This difference can be measured using the following energy
\begin{equation}
	\egrad = \sum_{(i,j)} \left\| 
	\left[
	\begin{matrix}
	s'_{i+1,j}-s'_{i,j}\\
	s'_{i,j+1}-s'_{i,j}\\
	\end{matrix}
	\right] \!\!-\!\!
	\left[
	\begin{matrix}
	I_{i+1,j}-I_{i,j}\\
	I_{i,j+1}-I_{i,j}\\
	\end{matrix} 
	\right]
	\right\|_2^2,
	\label{eq:GradientEnergy}
\end{equation}
where $\{I_{i,j}\}$ are intensity values from the input image, and
\begin{equation}
	s'_{i,j} = r_{i,j} \cdot \max(\sphcoef^T \bfH(\normalmap{i,j}), 0)
\end{equation}
are shading intensities for the normal map according to Eq.~\eqref{eq:sphericalharmoniclighting}, using the optimized albedo $\{r_{i,j}\}$ and spherical harmonic coefficients $\sphcoef$ from Sec.~\ref{sec:medium}. Minimizing the difference in gradients instead of intensities helps to attenuate the influence from illumination noises such as cast shadows, while preserving the features from the input image. \revise{Another benefit is that its optimality condition is a higher-order PDE that results in smoother solution and reduces unnatural sharp features~\cite{Lysaker2003}. One example is shown in Fig.~\ref{fig:shadow_cast}, where the formulation with gradient difference reduces the sharp creases around the nose and the mouth.
 (see Fig.~\ref{fig:shadow_cast})}.

Optimizing $\egrad{}$ alone is not sufficient for good results, since the problem is under-constrained. Thus we introduce two additional regularization terms for the normal map. First we note that the medium face model from Sec.~\ref{sec:medium} provides good approximation of the final shape. Thus we introduce the following energy to penalize the deviation between normal map and the normals from the medium face  
\begin{equation}
	\eclose  = \sum_{(i,j)} \| \normalmap{i,j} - \bfn_{i,j} \|_2^2,
\end{equation}
where $\bfn_{i,j}$ is computed from the medium face mesh according to Eq.~\eqref{eq:meshnormal}. In addition, we enforce smoothness of the normal map using an energy that penalizes its gradient
\begin{equation}
	\esmooth = \sum_{(i,j)} \| \normalmap{i+1,j} - \normalmap{i,j} \|_2^2 + \| \normalmap{i,j+1} - \normalmap{i,j} \|_2^2.
\end{equation}

Finally, we need to ensure the normal map is \emph{integrable}, i.e., given the normal map there exists a height field surface such that conditions~\eqref{eq:HeightFieldConditions} are satisfied. Note that if \eqref{eq:HeightFieldConditions} are satisfied, then $p_{i,j}$ and $q_{i,j}$ are the increments of function $z$ along the grid directions. Moreover, the total increment of $z$ along the close path that connects pixels $(i,j), (i+1, j), (i+1, j+1), (i, j+1)$ should be zero, which results in the condition
\begin{equation}
	p_{i,j} + q_{i+1, j} - p_{i, j+1} - q_{i,j} = 0.
	\label{eq:IngegrabilityCondition}
\end{equation}
For the normal map to be integrable, this condition should be satisfied at each pixel. Indeed, with condition~\eqref{eq:HeightFieldConditions} we can interpret $p$ and $q$ as partial derivatives $\frac{\partial z}{\partial u}, \frac{\partial z}{\partial v}$ where $u, v$ are the grid directions; then condition~\eqref{eq:IngegrabilityCondition} corresponds to $\frac{\partial p}{\partial v} = \frac{\partial q}{\partial u}$, which is the condition for $(p, q)$ to be a gradient field. We can then enforce the integrability condition using an energy
\begin{equation}
	\eint = \sum_{(i,j)} ( p_{i,j} + q_{i+1, j} - p_{i, j+1} - q_{i,j} )^2.
	\label{eq:IntegrabilityEnergy}
\end{equation}
Combining the above energies, we derive an optimization problem for computing the desirable normal map
\begin{equation}
	\min_{\mathbf{p}, \mathbf{q}} \quad \egrad + \weightnormalmap{1} \eclose + \weightnormalmap{2} \esmooth + \weightnormalmap{3} \eint,
	\label{eq:normalmapoptimization}
\end{equation}  
where the optimization variables $\mathbf{p}, \mathbf{q}$ are the values $\{p_{i,j}\}, \{ q_{i,j} \}$, and $\weightnormalmap{1}, \weightnormalmap{2}, \weightnormalmap{3}$ are user-specified weights. This nonlinear least-squares problem is again solved using the Levenberg-Marquardt algorithm.

Fig.~\ref{fig:enhance} shows a fine face model reconstructed using our method. Compared with the medium face model, it captures more geometric details and reduces the reconstruction error. Besides, it can be observed from the reconstruction results in last two columns that the initial coarse face model has a large influence on reconstruction accuracy.

\section{Experiments}
\label{sec:experiments}
This section presents experimental results, and compares our method with some existing approaches.

\begin{figure*}[!t]
	\centering
	\includegraphics[width=\textwidth]{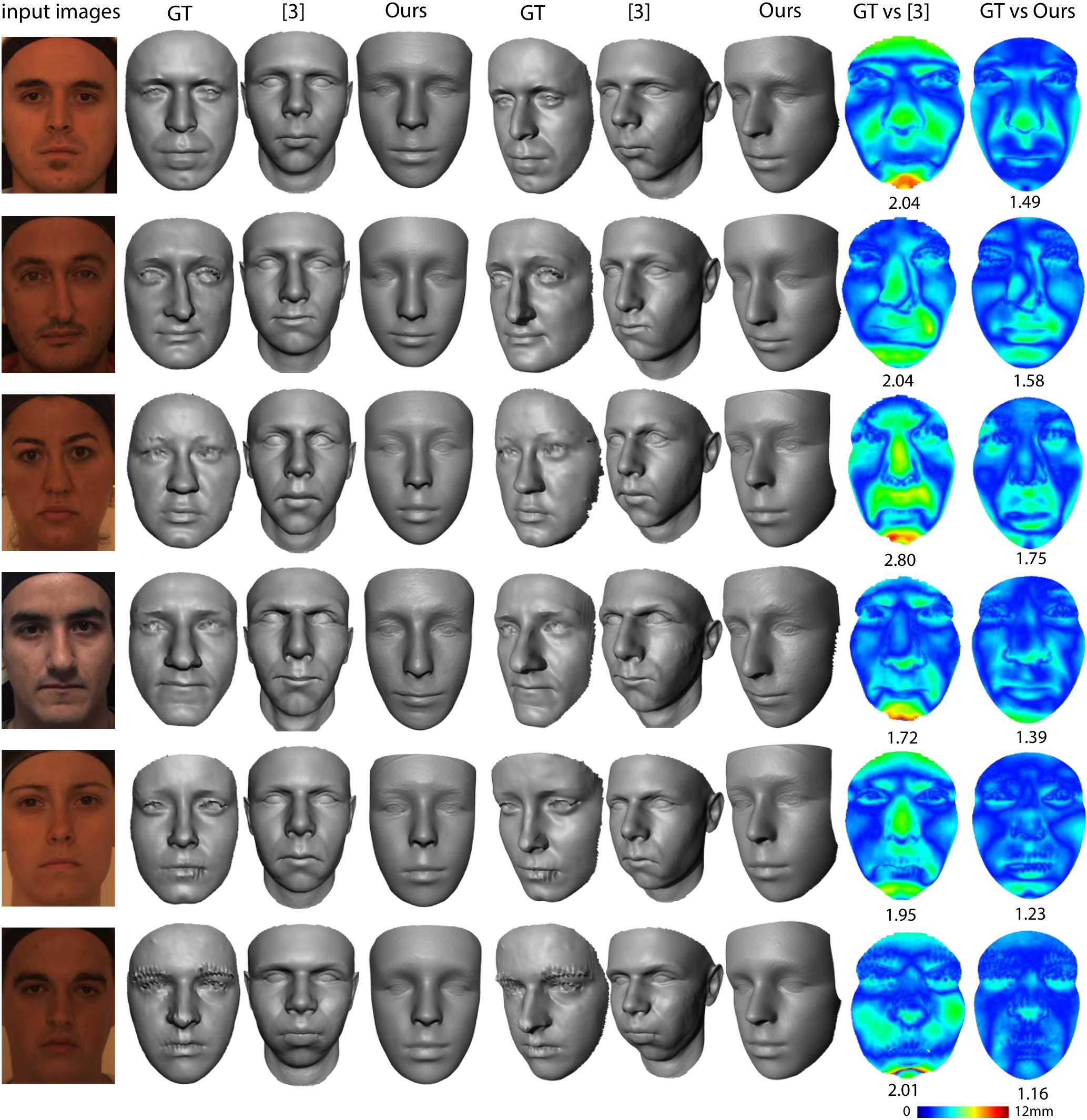}
	\caption{Facial reconstruction from images of frontal pose and neutral expression. For each input image, we show the ground truth (GT) as well as the results using out method and the method from~\cite{ZhuLYYL15}, each in two viewpoints. We also show the error maps (according to Eq.~\eqref{eq:ErrorMap}) for the two methods, together with their 3DRMSE.}
	\label{fig:exprement_A_8subjects}
\end{figure*}

\paraheading{Experimental setup}
To verify the effectiveness of our method, we tested it using the data set from the Bosphorus database~\cite{savran2008bosphorus}. This database provides structured-light scanned 3D face point clouds for 105 subjects, as well as their corresponding single-view 2D face photographs. For each subject, the database provides point clouds and images for different facial expressions and head poses. We ran our algorithm on the 2D images, and used the corresponding point clouds as ground truth to evaluate the reconstruction error. 55 subjects with low noises in their point clouds were chosen for testing. The reconstructed face is aligned with its corresponding ground truth face using iterative closest point (ICP) method~\cite{RusinkiewiczL01}. After alignment, we crop the face model at a radius of 85mm around the tip of the nose, and then compute the 3D Root Mean Square Error (3DRMSE):
\begin{equation}
	\sqrt{\sum_{i}(\bfX-\bfX^*)^2}/N,
	\label{eq:ErrorMap}
\end{equation}
where $\bfX$ is the reconstructed face, $\bfX^*$ is the grund truth, $N$ is the \revise{number of vertices} of the cropped frontal reconstructed face. We also computed the mean and standard deviation of all these errors.

\begin{figure*}[!t]
	\centering
	\includegraphics[width=0.9\textwidth]{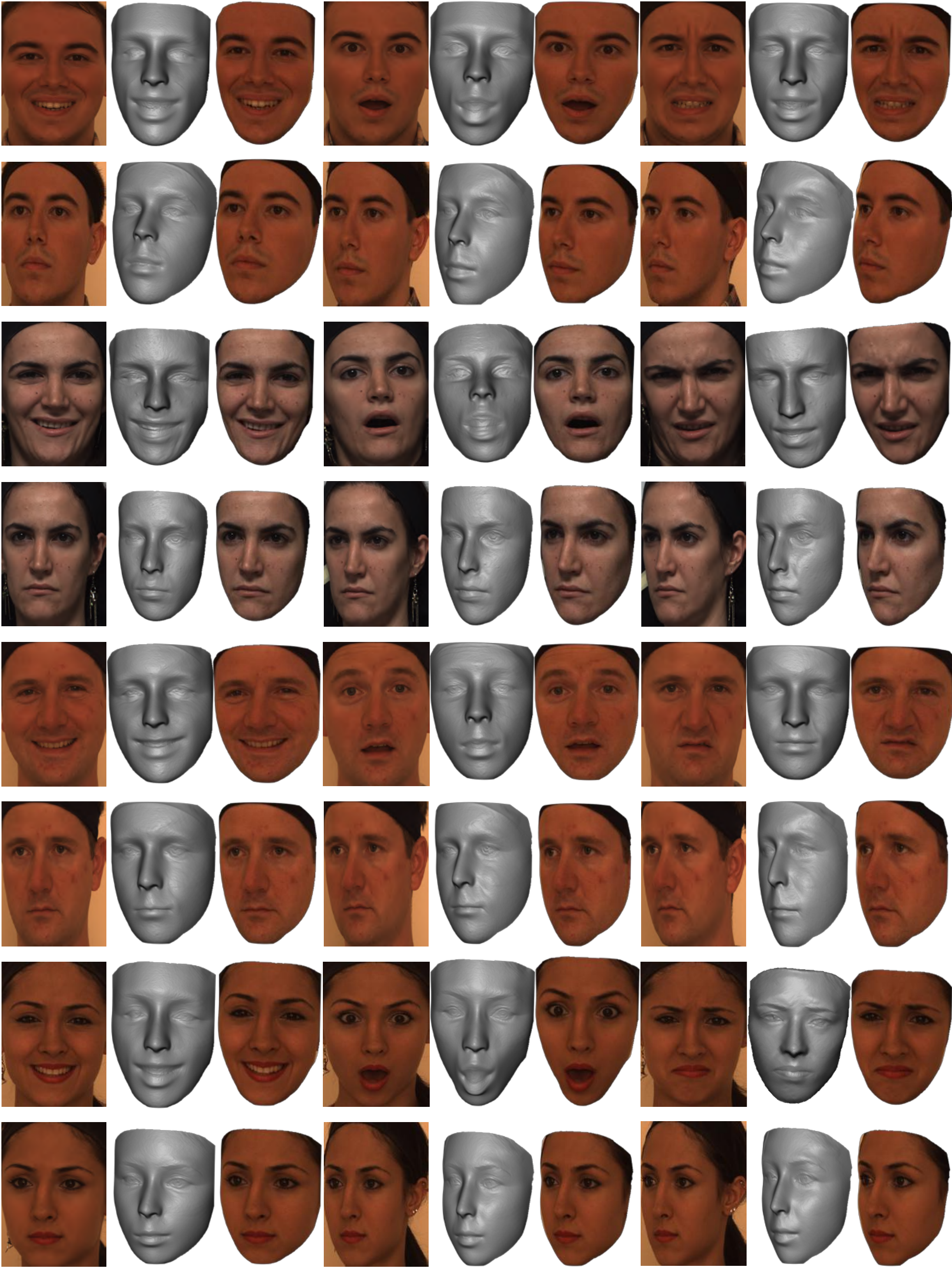}
	\caption{Face reconstructions of four subjects from images of frontal pose with different expressions (happy, surprise, disgust), and of different poses (Yaw $+10^{\circ}$, $+20^{\circ}$, $+30^{\circ}$) with neutral expression. For each input image, we show the reconstructed face mesh as well as its textured rendering.}
	\label{fig:exprement_B_4_subjects}
\end{figure*}

Our algorithm is implemented in C++ and is tested on a PC with an Intel Core i7-4710MQ 2.50 GHz CPU and 7.5 GB RAM. The weights in optimization problems \eqref{eq:2D3DFitting}, \eqref{eq:AlbedoLightingOptimization}, \eqref{eq:ReducedDeformationOptimization}, \eqref{eq:normalmapoptimization} are set as follows: $\weightfit{1} = \weightfit{2} = 1.5 \times 10^3$;$\weightalbedo{1} = 5$;$\weightsubspace = 20$; $\weightnormalmap{1} = 10, \weightnormalmap{2} = 10, \weightnormalmap{3} = 1$. The nonlinear least-squares problems are solved using the \ceres{} solver~\cite{ceres-solver}, with all derivatives evaluated using automatic differentiation. To speed up the algorithm, we downsample the high-resolution 2D images from the database to $30\%$ of their original dimensions before running our algorithm. The down-sampled images have about $400 \times 500$ pixels, for which the coarse, medium, and fine face construction steps take about 1 second, 2 minutes, and 1 minute respectively using our non-optimized implementation.
\begin{table}[tbp]
\centering
\caption{The mean and standard variation of our reconstructions for each pose and expression.}
\vspace{1.5mm}
\begin{tabular}{lccc}
\hline
Pose &Yaw $+10^{\circ}$ &Yaw $+20^{\circ}$ &Yaw $+30^{\circ}$\\ \hline      
3DRMSE &$1.73\pm0.33$ &$1.51\pm0.24$ &$1.44\pm0.32$\\ \hline
Expression &happy &surprise &disgust\\ \hline             
3DRMSE &$1.71\pm0.34$ &$2.05\pm0.49$ &$1.98\pm0.42$\\ \hline
\end{tabular}
\label{pose_exp}
\end{table}

\paraheading{Frontal and neutral faces}
We first tested our method on facial images of frontal pose and neutral expression, from 55 subjects in the Bosphorus database. For comparison we also ran the face reconstruction method from~\cite{ZhuLYYL15}, which is based on a 3DMM built from \bfm{} and \facewarehouse{}.
Fig.~\ref{fig:exprement_A_8subjects} presents the reconstruction results of six subjects using our method and~\cite{ZhuLYYL15}, and compares them with the ground truth faces. Thanks to the enhancement in the medium face step and the SFS recovery in the fine face step, our approach can not only obtain a more realistic global facial shape, but also accurately capture the person-specific geometric details such as wrinkles. Fig.~\ref{fig:exprement_A_8subjects} also shows the 3DRMSE for our results and the results using~\cite{ZhuLYYL15}. The mean and standard variation of 3DRMSE is $1.97\pm0.35$ for the results by method~\cite{ZhuLYYL15}, and $1.56\pm0.24$ for the results by our method. It can be seen that the mean error from our results are consistently lower than those from the method of~\cite{ZhuLYYL15}.
 \begin{figure*}[!t] 
	\centering
	\includegraphics[width=2\columnwidth]{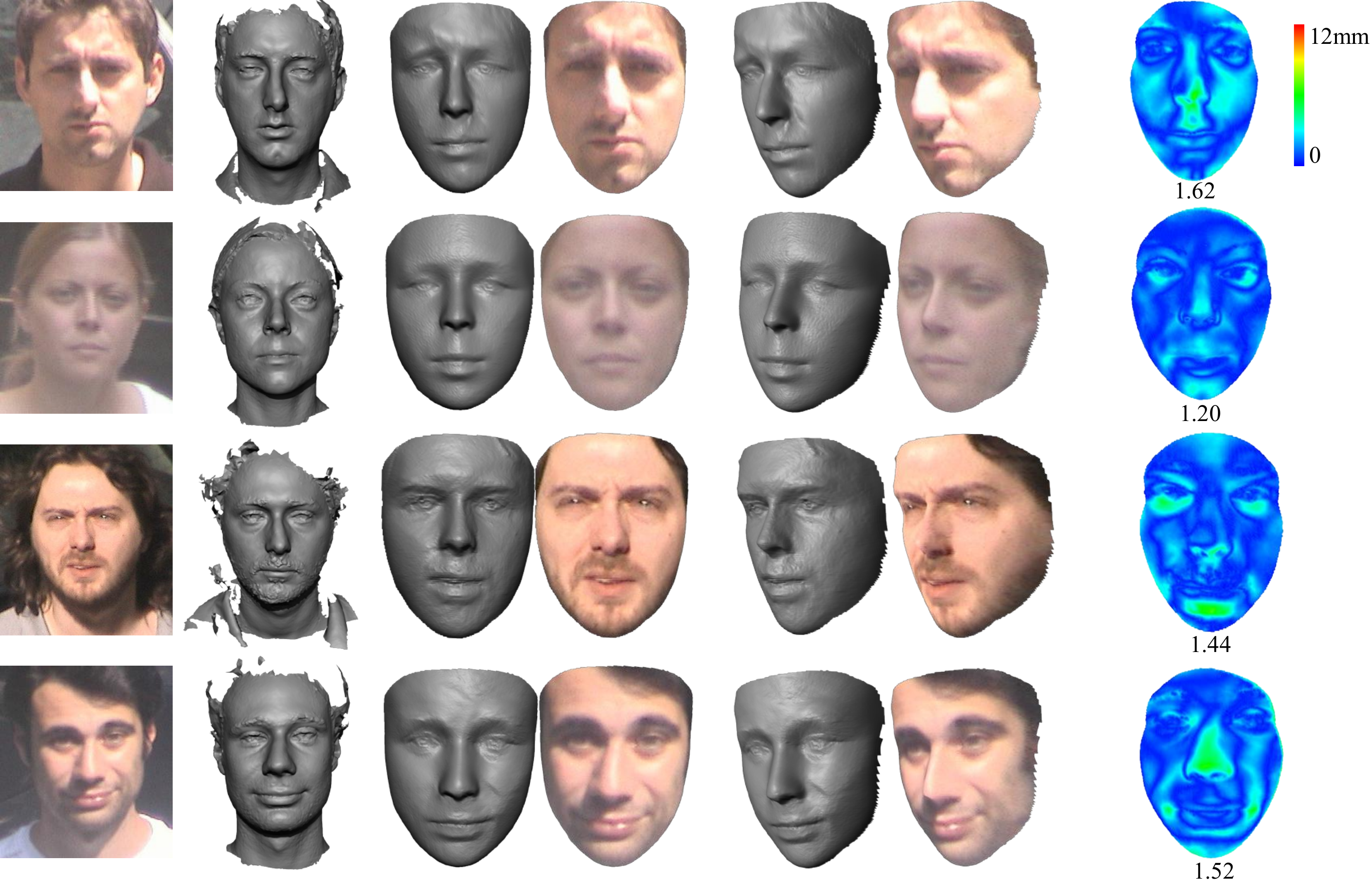}
	\caption{Face reconstructions of four subjects from the MICC dataset~\cite{Bagdanov2011} using our method. We show from left to right the input image, the ground truth, our \revise{reconstruction result (with texture)} in two view points, and error map (according to Eq.~\eqref{eq:ErrorMap}).}
	\label{fig:micc_results}
\end{figure*}

\revise{\paraheading{Near-frontal poses and expressions}}
We also tested our method on face images with \revise{near-frontal} poses and expressions. First, for each of the 55 subjects, we applied our method on their images of neutral expression with three types of poses: Yaw $+10^{\circ}$, $+20^{\circ}$, and $+30^{\circ}$. Then, we tested our approach on frontal faces with three non-neutral expressions: happy, surprise, and disgust. Among the 55 subjects, there are 25 of them with all three expressions present. We apply our method on these 25 subjects, and Table~\ref{pose_exp} shows the mean and standard deviation of 3DRMSE for each pose and expression. We can observe that the reconstruction results by our method are consistent for different poses and expressions, and the reconstruction errors are small. This is verified in Fig.~\ref{fig:exprement_B_4_subjects}, where we show the reconstruction results of four subjects under different poses and expressions.

\begin{figure}[t] 
   \centering
      \includegraphics[width=0.8\columnwidth]{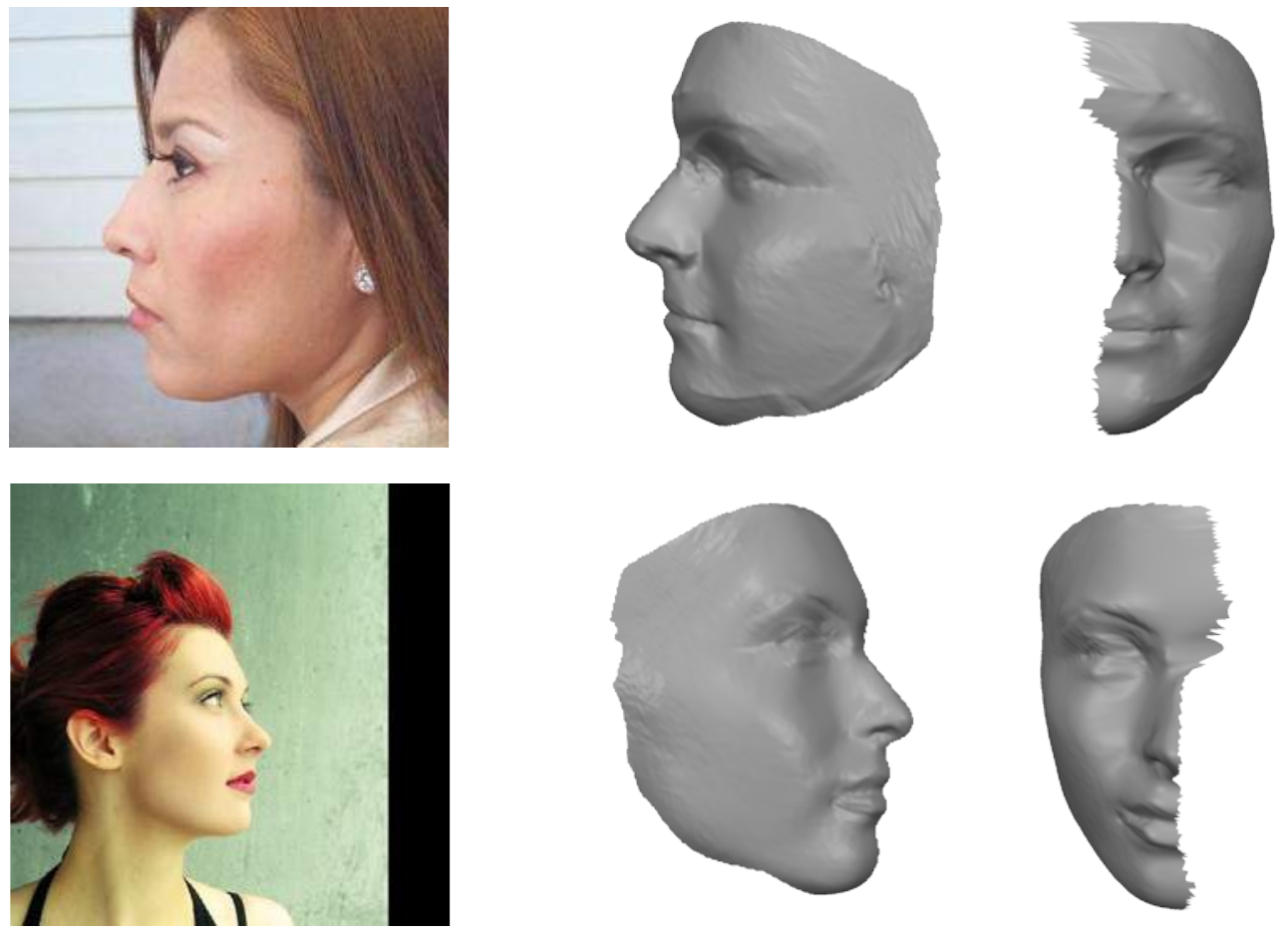}
   \caption{\revise{Face reconstructions of face images with very large pose using our method. We show from left to right the input image, and the reconstruction result from two viewpoints.}}
  \label{fig:bigpose}
 \end{figure}

Furthermore, using landmark detection methods designed for facial images with large pose (e.g., $90^\circ$), our approach can also reconstruct the 3D model well for such images. \revise{Two examples} are shown in Fig.~\ref{fig:bigpose}, where the landmarks are detected using the method from~\cite{zhu2016face}.

\paraheading{Unconstrained facial images}
To demonstrate the robustness of our approach on general unconstrained facial images, we compare our method with the \emph{structure from motion} (SFM) method~\cite{hernandez2017accurate} and the learning-based method~\cite{tran2016regressing} using the MICC dataset~\cite{Bagdanov2011}. The MICC dataset contains 53 video sequences of varying resolution, conditions and zoom levels for each subject, which is recorded in controlled, less controlled or uncontrolled environment. There is a structured-light scanning for each subject as the ground truth, and the reconstruction errors of the reconstruction results are computed following the way described in the above.
\begin{table}[tbp]
\centering
\caption{Quantitative results on the MICC dataset~\cite{Bagdanov2011}. The mean and standard variation of 3DRMSE, the runtimes.}
\vspace{1.5mm} 
\begin{tabular}{lccc}
\hline
Approach &3DRMSE &run time\\ \hline      
SFM~\cite{hernandez2017accurate} &$1.92\pm0.39$ &CPU 1min 13s\\
CNN-based methods~\cite{tran2016regressing} &$1.53\pm0.29$ &GPU 0.088s\\ 
Ours &$1.75\pm0.29$ &CPU 3min\\ \hline
\end{tabular}
\label{micc}
\end{table}
For each subject, we select the most frontal face image from the corresponding outdoor video and reconstruct the 3D face model by setting it as input. Table~\ref{micc} shows that our reconstruction error is close to~\cite{tran2016regressing} and lower than~\cite{hernandez2017accurate}. With the prior of reliable medium face and SFS recovery, our approach can also have good estimations on unconstrained \revise{images}. Fig.~\ref{fig:micc_results} presents the reconstruction results of four subjects using our method.

\begin{figure*}[!t] 
	\centering
	\includegraphics[height=0.95\textheight]{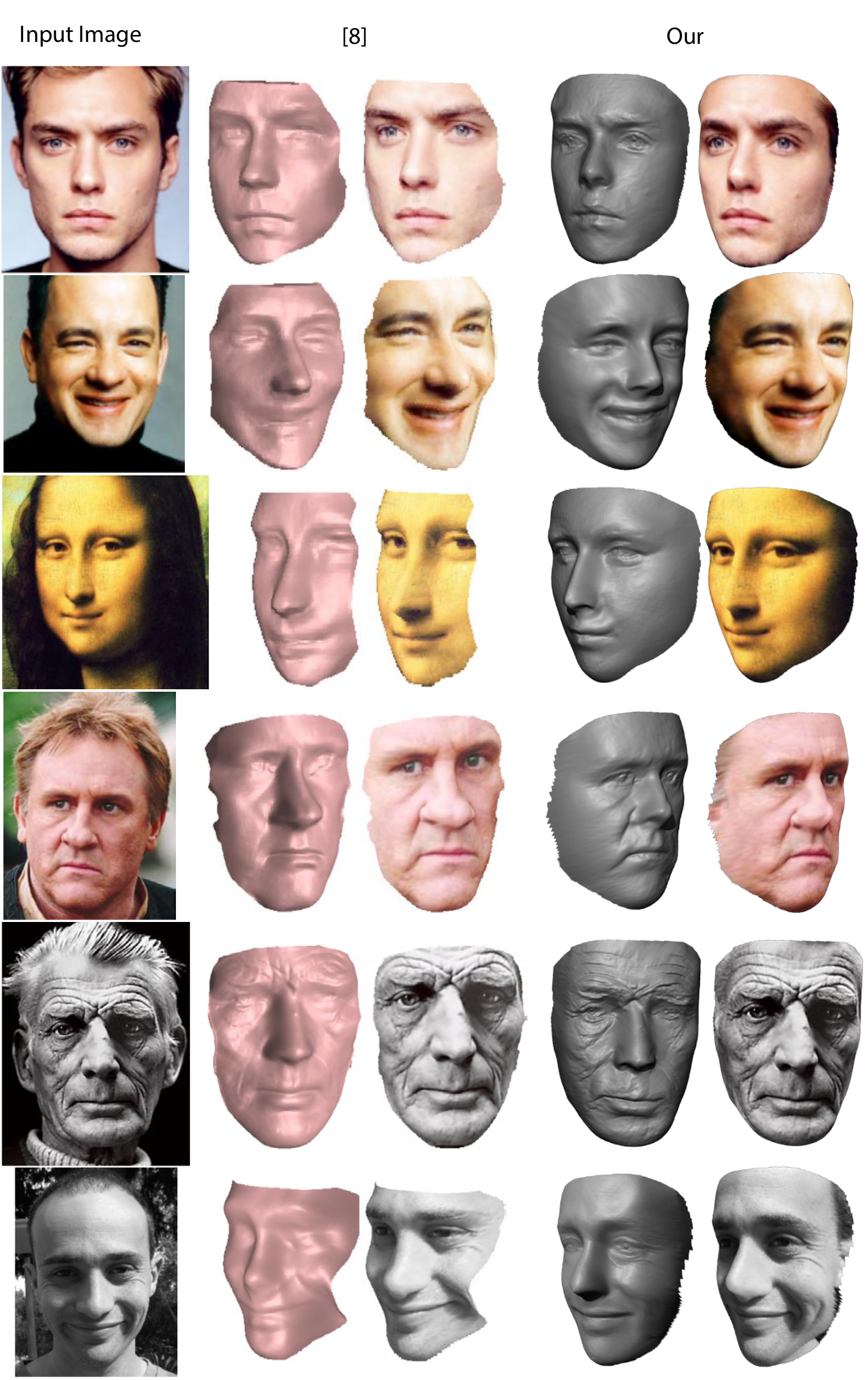}
	\caption{Face reconstructions from unconstrained images, using the method from~\cite{Kemelmacher-ShlizermanB11} and our method.}
	\label{fig:unconstrained}
\end{figure*}
We also compared our method with the SFS approach of~\cite{Kemelmacher-ShlizermanB11} on more general unconstrained facial images. 
Since there are no ground truth shapes for these images, we only compared them visually. For reliable comparison, we directly ran our algorithm on the example images provided in~\cite{Kemelmacher-ShlizermanB11}. Fig.~\ref{fig:unconstrained} presents the comparison results, showing both the reconstructed face geometry and its textured display. We can see that our approach produced more accurate reconstruction of the overall shape, and recovered more geometrical details such as winkles and teeth. Although both methods perform SFS reconstruction, there is major difference on how the shape and illumination priors are derived. In~\cite{Kemelmacher-ShlizermanB11} a reference face model is utilized as the shape prior to estimate illumination and initialize photometric normals; as the reference face model is not adapted to the target face shape, this can lead to unsatisfactory results. In comparison, with our method the medium face model is optimized to provide reliable estimates of the target shape and illumination, which enables more accurate reconstruction. 

\vspace{-2mm}
\section{Discussion and Conclusion}
\label{sec:Conclusion}

\revise{The main limitation of our method is that its performance for a given image depends on how well the overall face shape is covered by our constructed face model. This is because medium and fine face modeling have little effect on the coarse face shape; thus in order to achieve good results, the coarse face model needs to be close enough to the ground-truth overall shape, which can be achieved if the ground-truth face is close to the space spanned by our linear face model. By combining \facewarehouse{} and \bfm{} to construct the face model, our approach achieves good results on a large number of images. But for faces with large deviation from both \facewarehouse{} and \bfm{}, our method may not work well. One potential future work is to improve the face model by incorporating a larger variety of face datasets.}

\revise{Since we compute pixel values by multiplying albedo with lighting, there is an inherent ambiguity in determining albedo and lighting from given pixel values. Our approach alleviates the problem by using PCA albedo and second-order spherical harmonics lighting, but it does not fully resolve the ambiguity. Nevertheless, as we only intend to recover face geometry, such approach is sufficient for achieving good results.}

In this paper, we present a coarse-to-fine method to reconstruct a high-quality 3D face model from a single image. Our approach uses a bilinear face model and local corrective deformation fields to obtain a reliable initial face shape with large- and medium-scale features, which enables robust shape-from-shading reconstruction of fine facial details. The experiments demonstrate that our method can accurately reconstruct 3D face models from images with different poses and expressions, and recover the fine-scale geometrical details such as wrinkles and teeth. Our approach combines the benefits of low-dimensional face models and shape-from-shading, enabling more accurate and robust reconstruction. 

\vspace{-3mm}
\section*{Acknowledgments}
We would like to thank the reviewers for their time spent on reviewing our manuscript and their insightful comments helping us improving the article. This work was supported by the National Key R\&D Program of China (No. 2016YFC0800501), the National Natural Science Foundation of China (No. 61672481, No. 61672482 and No. 11626253), the Youth Innovation Promotion Association of CAS, and the One Hundred Talent Project of the Chinese Academy of Sciences.

\bibliographystyle{IEEEtran}
%%use following if all content of bibtex file should be shown
%\nocite{*}
\bibliography{SingleImageReconstruction}

\begin{IEEEbiography}[{\includegraphics[width=1in]{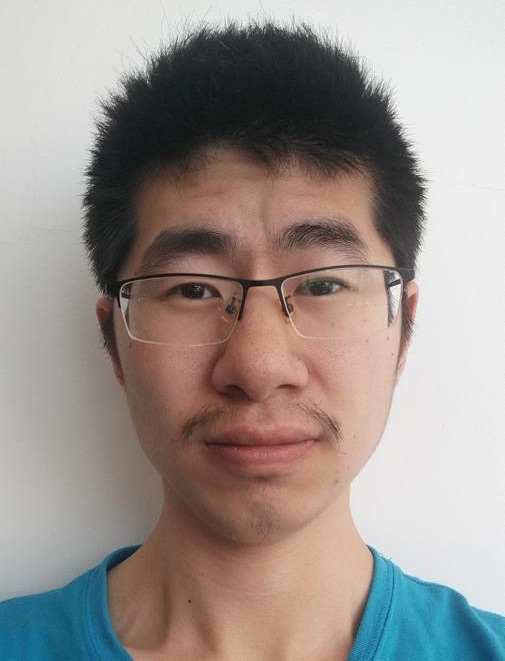}}]{Luo Jiang}
is currently working towards the PhD degree at the University of Science and Technology of China.
He obtained his bachelor degree in 2013 from the Huazhong University of Science and Technol-
ogy, China. His research interests include computer graphics, image processing and deep learning.
\end{IEEEbiography}

\vspace{-5mm}
\begin{IEEEbiography}[{\includegraphics[width=1in]{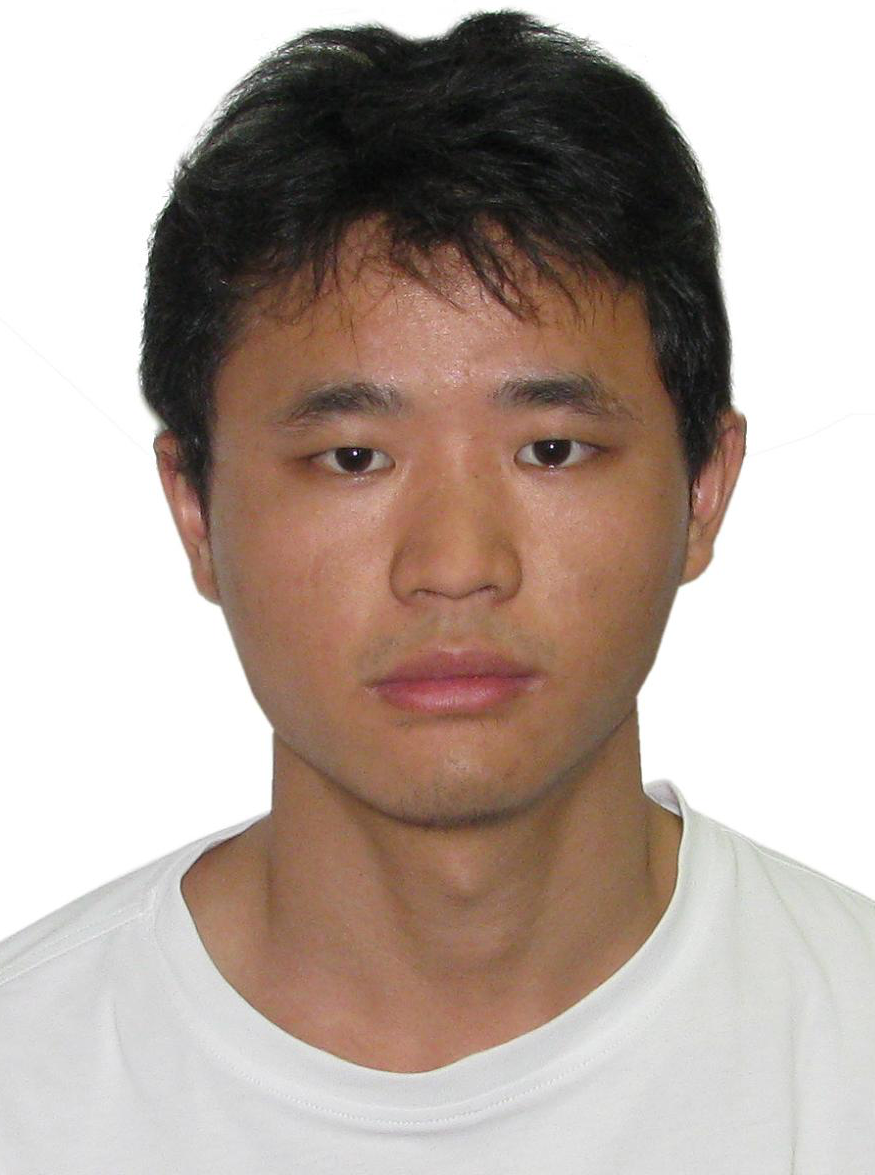}}]{Juyong Zhang}
is an associate professor in the School of Mathematical Sciences at University of Science and Technology of China. He received the BS degree from the University of Science and Technology of China in 2006, and the PhD degree from Nanyang Technological University, Singapore. His research interests include computer graphics, computer vision, and numerical optimization. He is an associate editor of The Visual Computer.
\end{IEEEbiography}

\vspace{-5mm}
\begin{IEEEbiography}[{\includegraphics[width=1in]{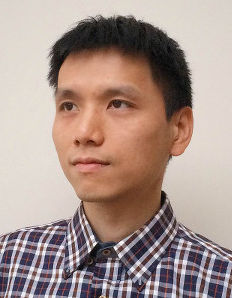}}]{Bailin Deng}
is a lecturer in the School of Computer Science and Informatics at Cardiff University. He received the BEng degree in computer software (2005) and the MSc degree in computer science (2008) from Tsinghua University (China), and the PhD degree in technical mathematics from Vienna University of Technology (Austria). His research interests include geometry processing, numerical optimization, computational design, and digital fabrication. He is a member of the IEEE.
\end{IEEEbiography}

\vspace{-5mm} \begin{IEEEbiography}[{\includegraphics[width=1in]{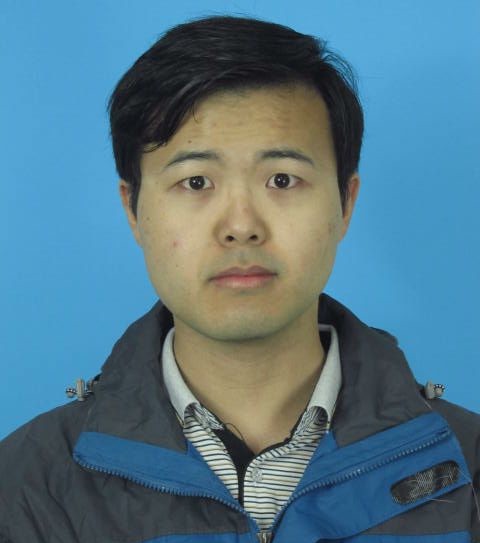}}]{Hao Li}
received the BSc degree in 2011 from the University of Science and Technology of China. His research interests include computer graphics and image processing.
\end{IEEEbiography}

\vspace{-10mm}
\begin{IEEEbiography}[{\includegraphics[width=1in]{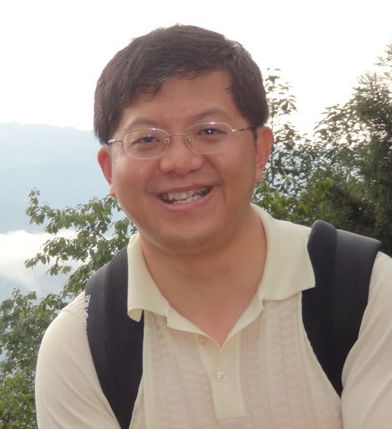}}]{Ligang Liu}
is a Professor at the School of Mathematical Sciences, University of Science and Technology of China. His research interests include digital geometric processing, computer graphics, and image processing. He serves as the associated editors for journals of IEEE Transactions on Visualization and Computer Graphics, IEEE Computer Graphics and Applications, Computer Graphics Forum, Computer Aided Geometric Design, and The Visual Computer. He served as the conference co-chair of GMP 2017 and the program co-chairs of GMP 2018, CAD/Graphics 2017, CVM 2016, SGP 2015, and SPM 2014. His research works could be found at his research website: http://staff.ustc.edu.cn/lgliu.
\end{IEEEbiography}

\end{document}